\documentclass[10pt, a4paper]{article}
\usepackage[final]{lrec-coling2024}

\usepackage{hyperref}
\usepackage{latexsym}
\usepackage{danudefs}
\usepackage{algorithm}
\usepackage{color}
\usepackage{booktabs}
\usepackage{xspace}
\usepackage{url}
\usepackage[export]{adjustbox}
\usepackage{amssymb}
\usepackage{amsmath}
\usepackage{subfigure}
\usepackage{float}

\usepackage{inconsolata}
\usepackage{microtype}

\makeatletter
\renewcommand{\@thesubfigure}{\hskip\subfiglabelskip}
\makeatother

\usepackage[english]{babel}
\usepackage{hyperref}
\addto\captionsenglish{%
}
\addto\extrasenglish{%
}

\title{Evaluating Unsupervised Dimensionality Reduction Methods for Pretrained Sentence Embeddings}

\name{Gaifan Zhang\thanks{Work done while the first author was a student at 
the University of Liverpool.}$^1$, Yi Zhou$^2$, Danushka Bollegala$^3$} 

\address{Columbia University$^1$, Cardiff University$^2$,  University of Liverpool$^3$ \\
         zgaifan@gmail.com, zhouy131@cardiff.ac.uk, danushka@liverpool.ac.uk\\}

\abstract{
Sentence embeddings produced by Pretrained Language Models (PLMs) have received wide attention from the NLP community due to their superior performance when representing texts in numerous downstream applications.
However, the high dimensionality of the sentence embeddings produced by PLMs is problematic when representing large numbers of sentences in memory- or compute-constrained devices.
As a solution, we evaluate unsupervised dimensionality reduction methods to reduce the dimensionality of sentence embeddings produced by PLMs.
Our experimental results show that simple methods such as Principal Component Analysis (PCA) can reduce the dimensionality of sentence embeddings by almost $50\%$, without incurring a significant loss in performance in multiple downstream tasks.
Surprisingly, reducing the dimensionality further \emph{improves} performance over the original high dimensional versions for the sentence embeddings produced by some PLMs in some tasks.
 \\ \newline  }

\begin{document}
\maketitleabstract

\section{Introduction}
\label{sec:intro}


Sentence embedding models represent a given input sentence using a fixed dimensional vector, which is independent of the length of the input sentence~\cite{reimers-gurevych-2019-sentence,gao-etal-2021-simcse}.
Sentence embeddings have significantly improved performance in numerous downstream NLP tasks such as information retrieval~\cite{Weize:2022,Palangi:2016}, question answering~\cite{Hao:2019}, and machine translation~\cite{wang-etal-2017-sentence} to name a few~\cite{Choi:2020}.
However, compared to static word embeddings~\cite{Glove,Milkov:2013}, which typically have a smaller number of dimensions (ca. 50-300), sentence embeddings produced by PLMs are usually high dimensional (ca. 1024-4096).
This is problematic due to several reasons as described next.

First, storing pre-computed sentence embeddings requires larger memory/disk space.
For example, in dense retrieval systems~\cite{Weize:2022,SPS}, documents must be pre-embedded and stored in order to efficiently process queries at retrieval time.
Therefore, the storage requirements increase linearly with the number of documents in the collection.
Although it is efficient to store all the embeddings in memory to reduce the search latency as opposed to reading from a disk-based storage, this is not possible due to the high dimensionality of the sentence/document embeddings.
Moreover, training deep learning models with high-dimensional sentence embeddings is problematic because GPUs have limited memory buffers.
This requires carefully selecting training data batches and results in latency overheads when transferring sentence embeddings back-and-forth between GPUs.

Second, the computation time of the inner-products between two sentence embeddings increases linearly with the dimensionality of the embedding.
For example, in dense retrieval, we must compute the inner-product (or equivalently cosine similarity if the embeddings are $\ell_2$ normalised) between the query embedding and each of the document embeddings to rank the documents that are relevant to a query~\cite{pmlr-v162-menon22a,SPS}.
This is prohibitively expensive when a large number of documents must be ranked within millisecond order retrieval times.  

Given this trade-off between the dimensionality and the accuracy of sentence embeddings, in this paper, we consider the following question: \textbf{Can we reduce the dimensionality of pre-computed sentence embeddings without significantly sacrificing the performance in downstream tasks that use those \emph{dimensionality-reduced} sentence embeddings?}
Although a diverse set of dimensionality reduction methods have been proposed, 
from a real-world large-scale application perspective, dimensionality reduction methods that satisfy the following properties are desirable:\\
(a) In contrast to training lower-dimensional sentence embeddings from scratch or training lower-dimensional student models via distillation~\cite{anil2018large}, we would prefer dimensionality reduction methods that are \emph{computationally lightweight} such that they can be applied in a \emph{post-processing} stage to reduce the dimensionality of pre-computed sentence embeddings. 
Distillation-based methods require additional training data for training the student model, as well as a larger memory to load the teacher model. 
Moreover, the inference time required by the student model, which will be used after distillation must also be taken into account. 
On the other hand, dimensionality reduction methods are particularly desirable when we have a large and continuously growing number of sentences as in the case of dense information retrieval.\\
(b) We would prefer \emph{unsupervised} dimensionality reduction methods over supervised ones such that no labelled data are required for a specific downstream task~\cite{zhao2022compressing}. 
Such labelled instances might not be available in specialised domains and in larger quantities.
This helps us to produce dimensionality-reduced sentence embeddings that are independent of a particular task, thus more likely to generalise well to multiple different tasks.

We present a novel analysis of unsupervised dimensionality reduction methods for sentence embeddings: truncated Singular Value Decomposition~\cite[\textbf{SVD};][]{MatrixCookbook}, Principal Component Analysis~\cite[\textbf{PCA};][]{andrews2016compressing}, Kernel PCA~\cite[\textbf{KPCA};][]{kernelPCA}, Gaussian Random Projections~\cite[\textbf{GRP};][]{bingham2001random} and Autoencoders~\cite{Vincent:ICML:2008}. 
Lower-dimensional projections can be learnt in an \emph{inductive} setting (uses only the \emph{train} sentences to learn the projection), or in a \emph{transductive} setting (uses unlabelled \emph{test} sentences in addition to the train sentences).
The inductive setting is desirable in scenarios where we have a continuous stream of test sentences such as in dense retrieval, whereas the transductive setting is sufficient when the test sentences are fixed and known in advance during projection learning time.

We use six popular sentence encoders in three tasks: semantic textual similarity (STS) measurement~\cite{cer2017semeval}, entailment prediction~\cite{marelli-etal-2014-sick}, and TREC question-type classification~\cite{hovy-etal-2001-toward}. 
In particular, semantic textual similarity is a popular NLP task that is frequently used to measure the accuracy of sentence embeddings~\cite{cer2017semeval}.
The ability to recognise textual entailment is also considered a fundamental task for evaluating natural language understanding~\cite{dzikovska-etal-2013-semeval,Yokote:AAAI:2012}. 
Moreover, question classification is a sentence classification task, which requires an accurate sentence representation to obtain good accuracy~\cite{li-roth-2002-learning}.
Overall, PCA proves to be the most effective method for sentence embedding compression. 
Previous research has demonstrated the effectiveness of PCA for various compression tasks~\cite{raunak2019effective, reddy2020analysis}, but we are the first to conduct a systematic study specifically for sentence embedding compression.

Our experimental results show that in both transductive as well as inductive settings, we can use PCA to reduce the dimensionality of diverse sentence embeddings by ca. $50\%$ within a $1\%$ loss in performance in the target tasks.
Interestingly, reducing dimensionality \emph{improves} accuracy for some sentence encoders such as \textit{all-mpnet-base-v2} for STS and \textit{msmarco-roberta-base-v2} for TREC.

\section{Related Work}
\label{sec:related}

\paragraph{Neural Network Compression:}
The majority of work for compression focuses on learning neural network models with fewer parameters.
Different techniques have been proposed for this purpose such as pruning (neuron/layer/weight dropout)~\cite{Han:2015,Li:2017,Lee:2019}, quantization~\cite{Han:2016,chen-EtAl:2016:P16-11}, distillation~\cite{jiao-etal-2020-tinybert,DistilBERT}, etc.
However, in this paper we \emph{do not} consider the problem of compressing PLMs, but focus only on the dimensionality reduction of the sentence embeddings produced by a given sentence encoder.

\paragraph{Word Embedding Compression:}
Several prior work has considered the problem of compressing pre-trained static word embeddings.
\citet{andrews2016compressing} adds sparsity and non-negative encoding to a specific autoencoder scheme, thereby compressing original dense vector embeddings. 
\citet{kim-etal-2020-adaptive} propose an adaptive compression method that uses code-book to represent words as discrete codes of different lengths. 
To reduce the number of parameters, \citet{shu2018compressing} assign a small set of basis vectors to each word, with the storage efficiency maximised by a composition coding approach. 
However, our focus in this paper is sentence embeddings and \emph{not} word embeddings.

\paragraph{Sentence Embedding Compression:}
Compared to model/word embedding compression work described above, lower-dimensional sentence embeddings are understudied.
\citet{zhao2022compressing} proposed homomorphic projective distillation to compress sentence embeddings 
using labelled training instances for textual entailment to learn a projection layer for a transformer-based encoder.
In contrast, the dimensionality reduction methods that we evaluate are all \emph{unsupervised}, thus not requiring labelled data.
Moreover, the methods we consider do not require student network training, which can be computationally expensive.

\section{Dimensionality Reduction Methods}
\label{sec:dim-reduct}


Let us assume that we are given a sentence embedding model, $M$, which returns a $d$-dimensional embedding, $\vec{s} (\in \R^d)$ to a sentence $s$.
Given a set of (train) sentences, $\cD_{\rm train}$, we will learn a $d'$-dimensional projection ($d' < d$), $f: \R^d \rightarrow \R^{d'}$ that would project a given sentence embedding.
We consider SVD, PCA, GRP, KPCA and Autoencoders as the unsupervised dimensionality reduction methods for the purpose of learning this projection.\footnote{Details of each method are in Appendix}
These methods have been selected due to their superior performance and popular applications.
For example, PCA has been used for compressing word embeddings~\cite{raunak2019effective}, 
KPCA has been used for feature extraction~\cite{ayesha2020overview, gupta-etal-2019-improving},
SVD has shown excellent performance in diverse word embedding-related tasks~\cite{levy2015improving},
random projection is a popular lightweight lower-dimensional projection method~\cite{schmidt2018stable}, and
autoencoders have been used to learn embeddings~\cite{socher2011dynamic,chandar2014autoencoder}.
In our preliminary experiments, we observed that an autoencoder with a single hidden layer was producing comparable performance to ones with more than two hidden layers, despite the former being fast to train and infer with.
Therefore, in the remainder of our experiments, we use autoencoders with a single hidden layer.
The learnt $f$ is then used to project test sentences, $\cD_{\rm test}$, to a $d'$-dimensional space.


\begin{figure*}[t]
    \centering
    \subfigure[Transductive setting]{
        \rotatebox{90}{\footnotesize{~~~~~~~~~~~~~~~~~~~~~~~SICK-E~~~~~~~~~~~~~~~~~~~~~~~~~~~~~~~~~~~~~~~~~~~~~TREC ~~~~~~~~~~~~~~~~~~~~~~~~~~~~~~~~~~~~~~~~~~~~~STS-B}}
        \begin{minipage}[b]{0.4\linewidth}
        \includegraphics[width=1.1\linewidth,trim=0 0 0 28,clip]{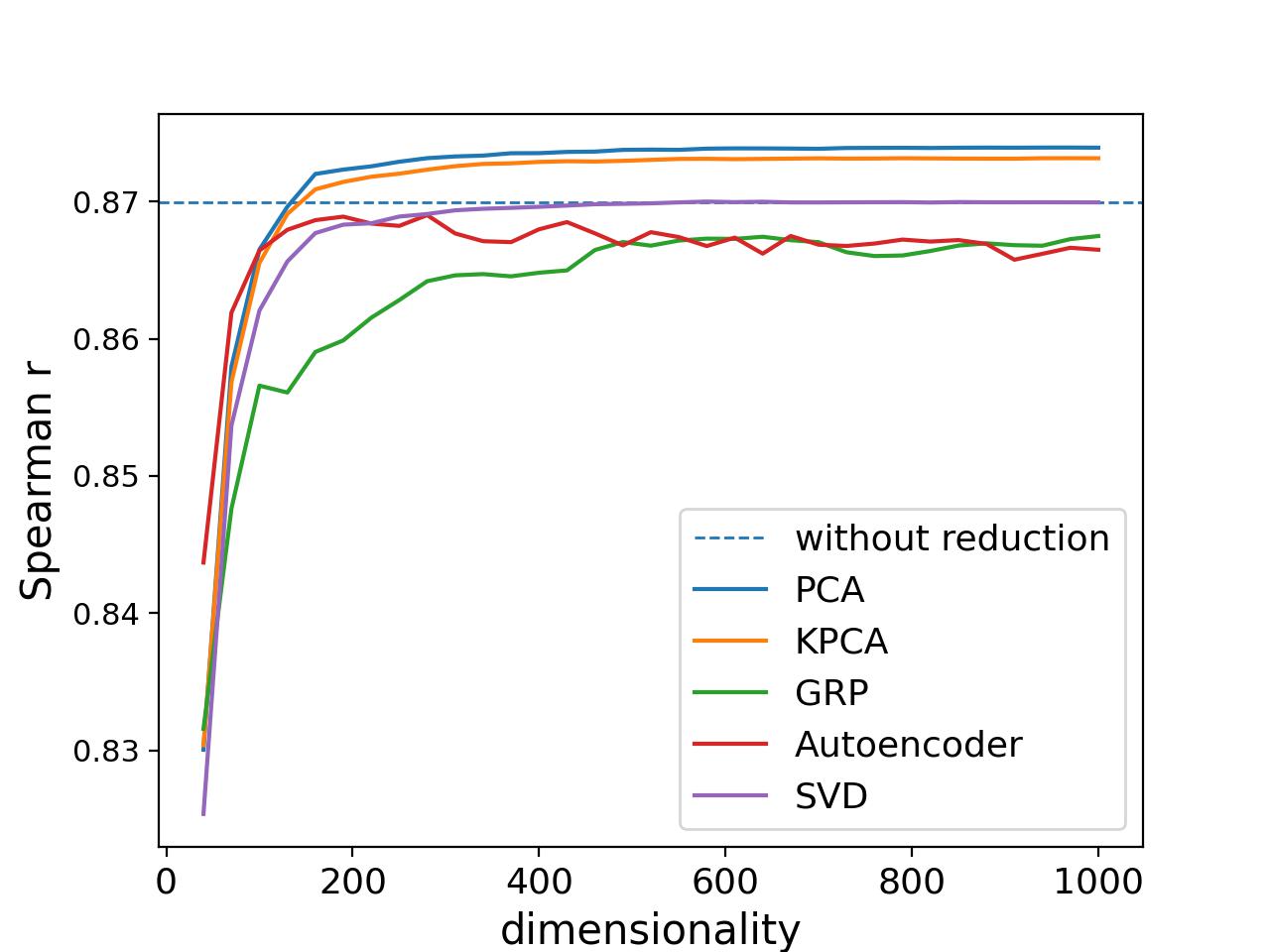}
        \includegraphics[width=1.1\linewidth,trim=0 0 0 28,clip]{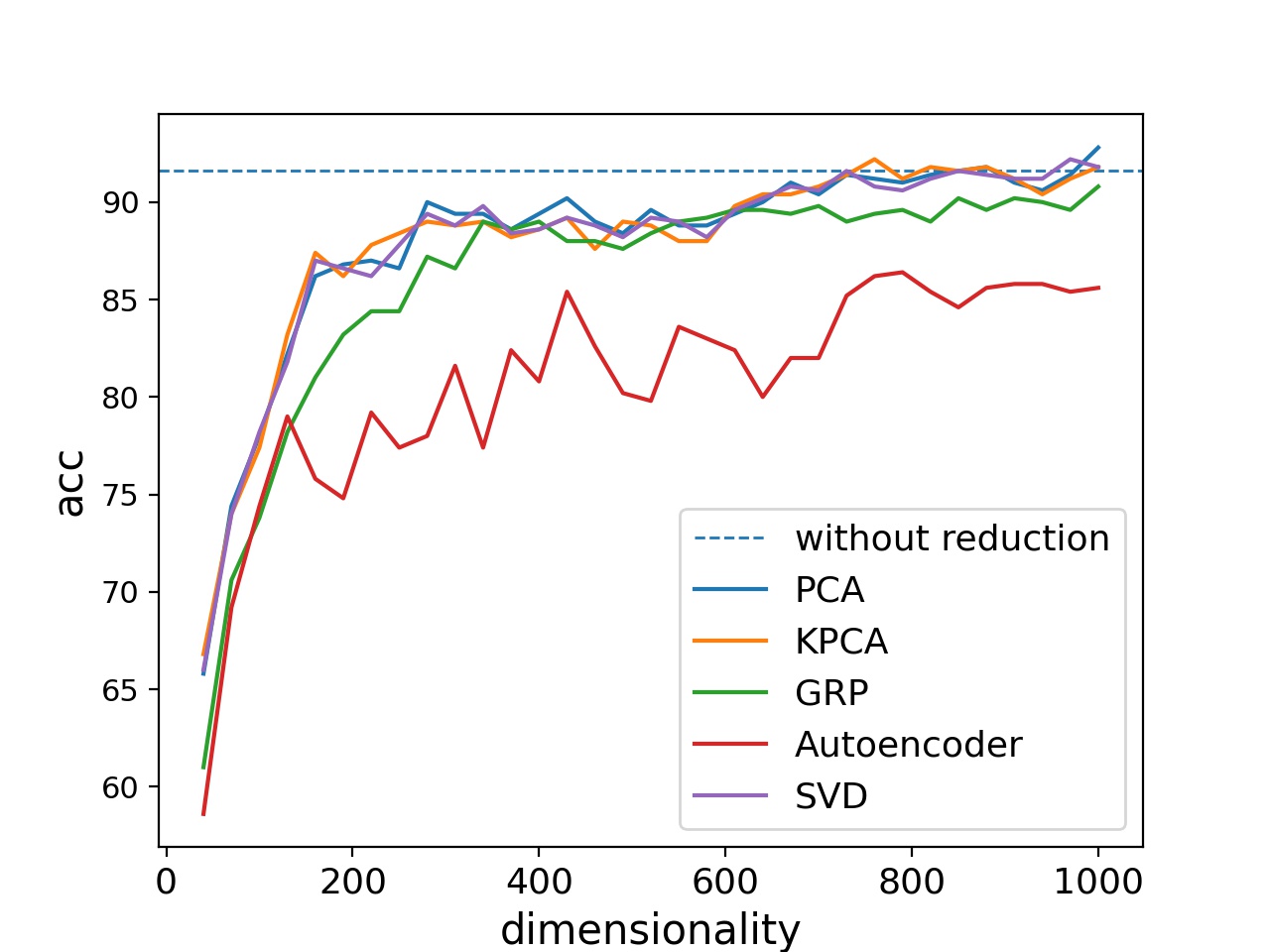}
        \includegraphics[width=1.1\linewidth,trim=0 0 0 28,clip]{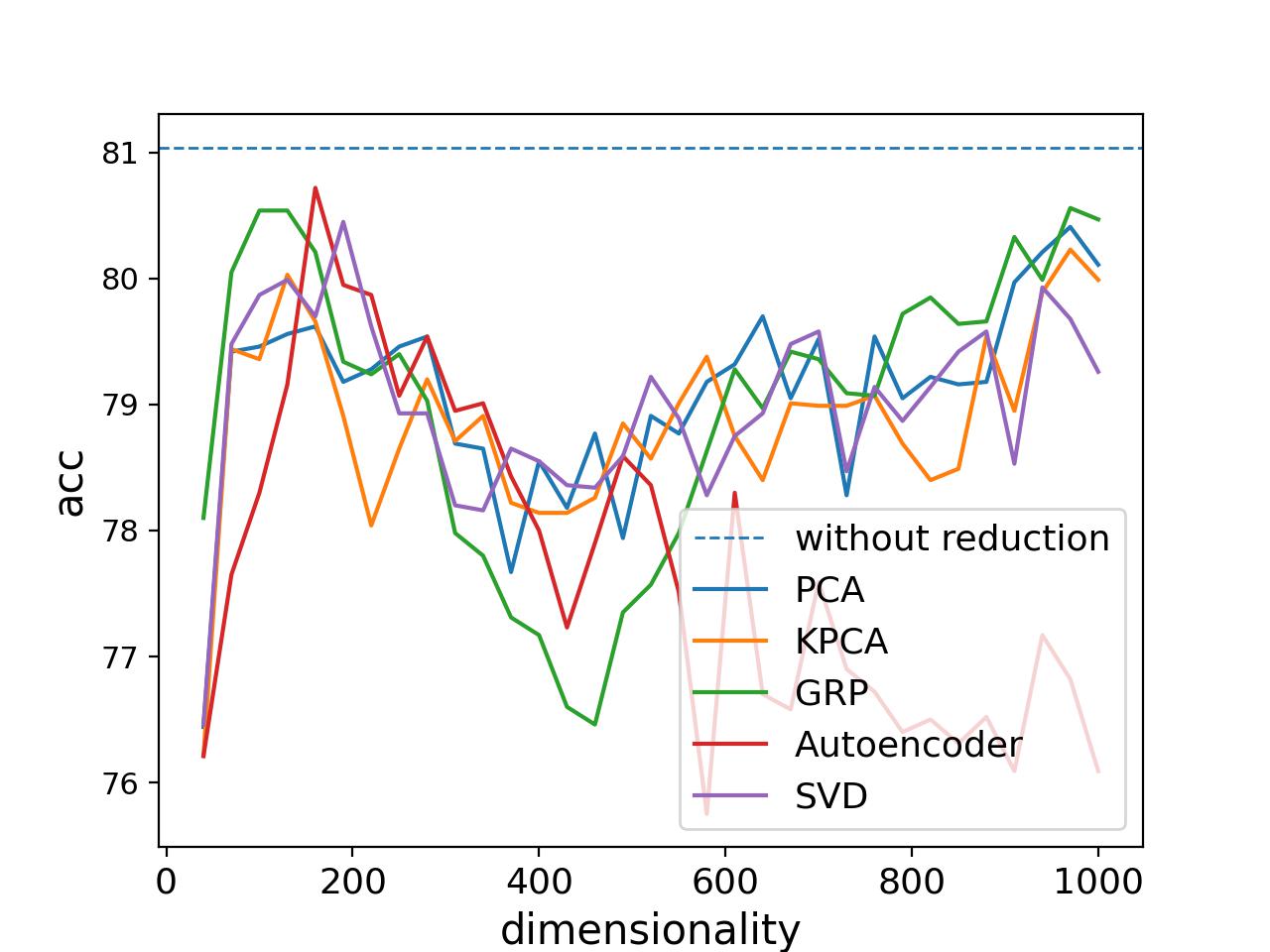}
        \end{minipage}
    }
    \subfigure[Inductive setting]{
        \begin{minipage}[b]{0.4\linewidth} 
        \includegraphics[width=1.1\linewidth,trim=0 0 0 28,clip]{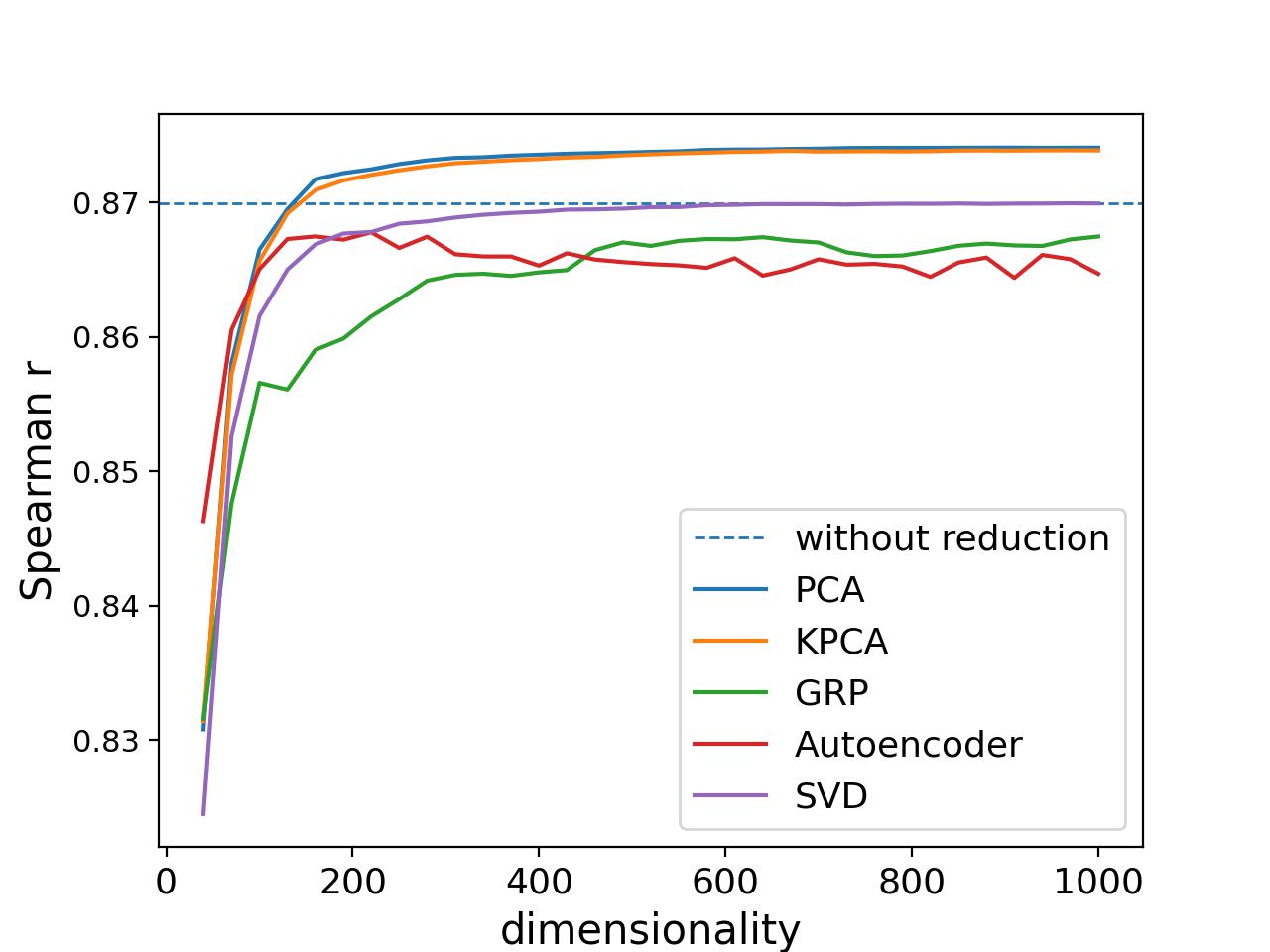}
        \includegraphics[width=1.1\linewidth,trim=0 0 0 28,clip]{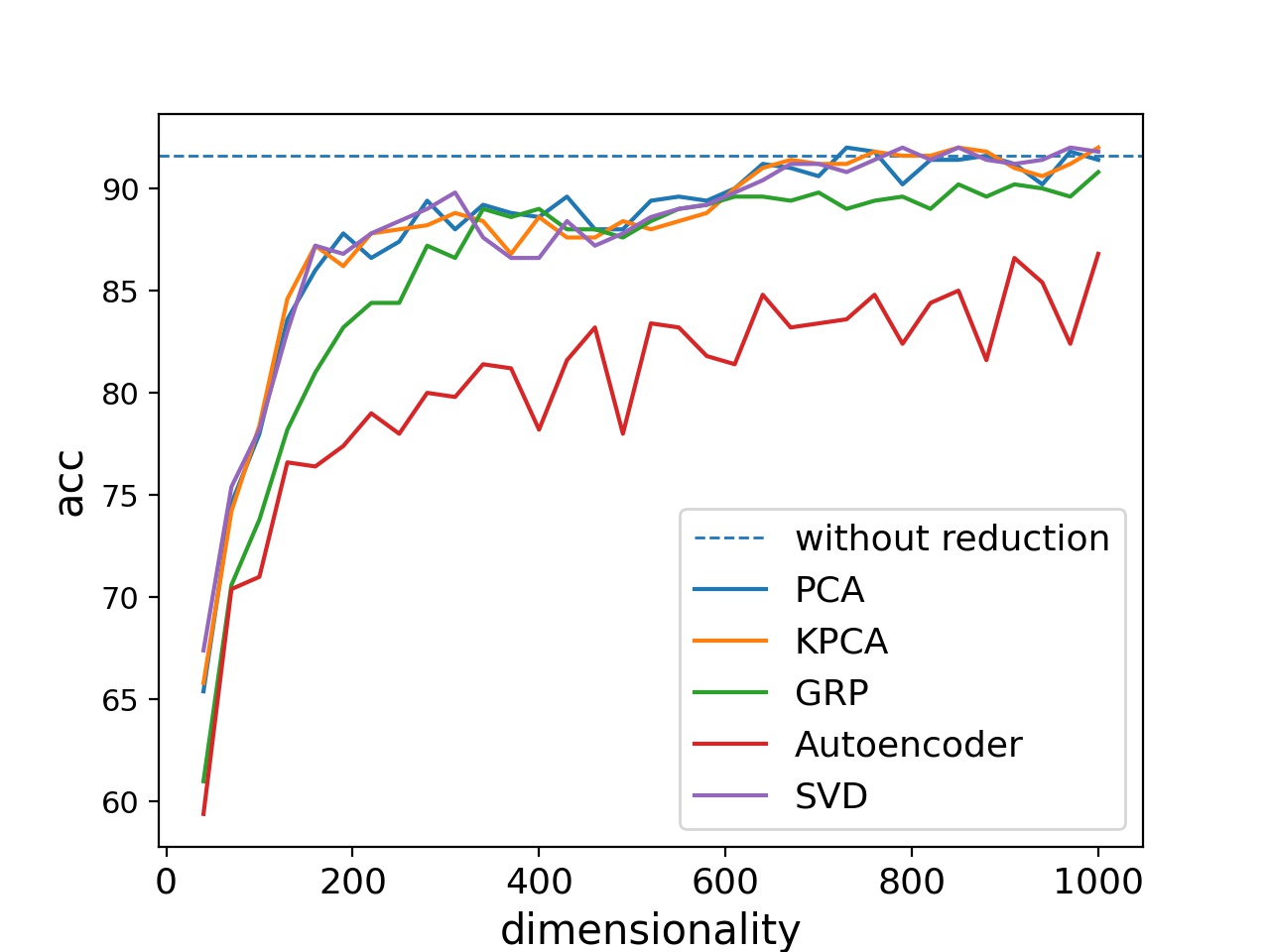}
        \includegraphics[width=1.1\linewidth,trim=0 0 0 28,clip]{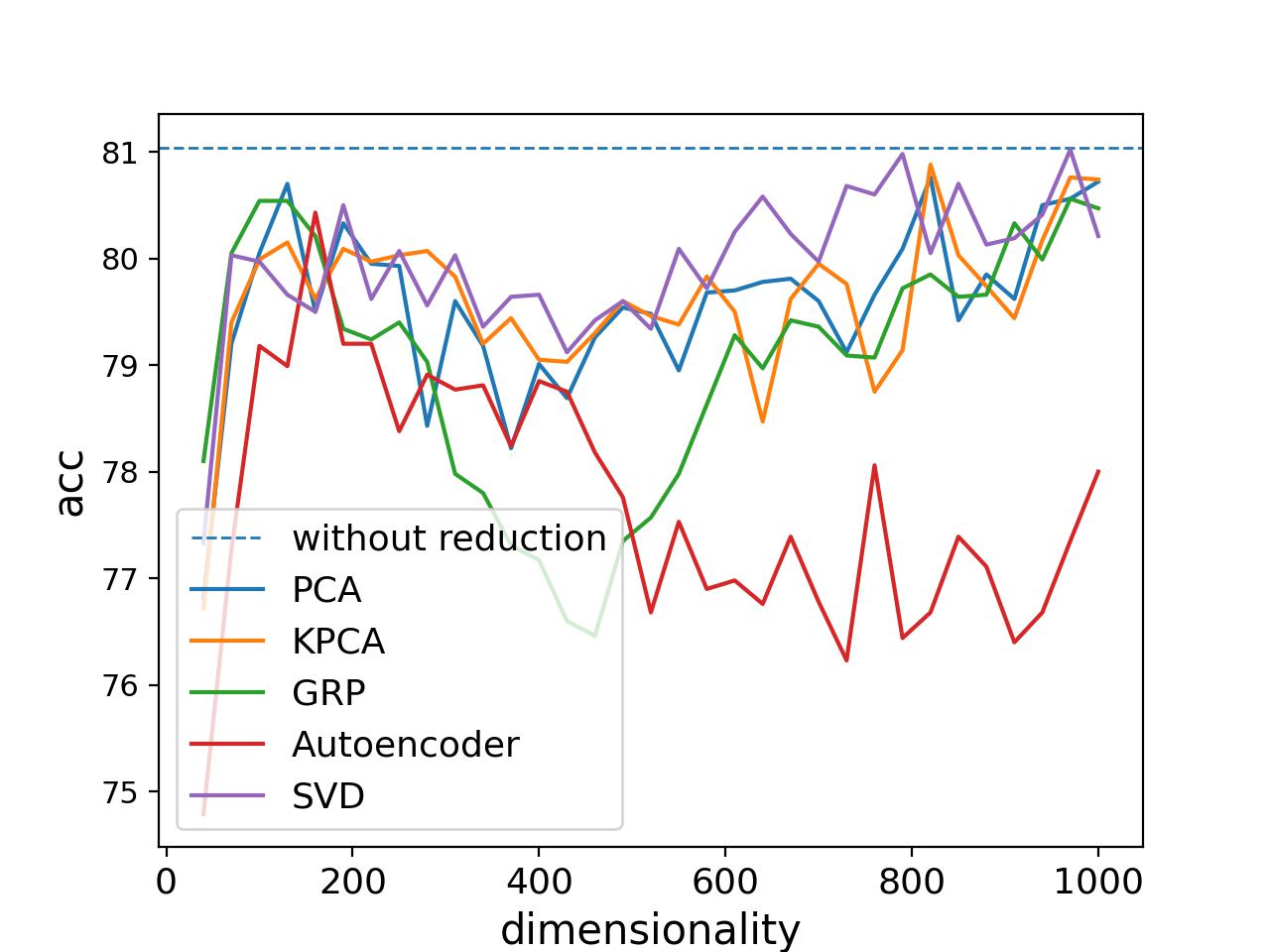}
        \end{minipage}
    }
    \caption{Performance of the original \textit{sup-simcse-roberta-large} sentence embeddings and its dimensionality reduced versions produced using different methods on STS-B (top), TREC (middle) and SICK-E (bottom) datasets. Results for the transductive and inductive settings are shown respectively on the left and right.}
    \label{fig:main-results}
\end{figure*}

\section{Experiments}




To evaluate the efficiency and effectiveness of the dimensionality reduction methods described in \autoref{sec:dim-reduct}, we apply them to reduce the dimensionality of sentence embeddings which are produced by six transformer PLMs\footnote{All models are available at \url{huggingface.co/sentence-transformers} and \url{https://huggingface.co/princeton-nlp}}: \textit{all-mpnet-base-v2} (\textbf{mpnet}),
\textit{stsb-bert-base} (\textbf{sbert-b}), \textit{msmarco-roberta-base-v2} (\textbf{roberta}), \textit{paraphrase-xlm-r-multilingual-v1} (\textbf{xml-r}),  \textit{stsb-bert-large} (\textbf{sbert-l}), and \textit{sup-simcse-roberta-large} (\textbf{simcse}).
We evaluate using three datasets: STS-B for STS, TREC for question-type classification, and SICK-E for entailment prediction.
We use an NVIDIA RTX A6000 GPU. The scikit-learn (0.24.2) is used for SVD, PCA, KPCA and GRP, and Keras (2.2.4) is used for the autoencoder. 
We fix these settings in all experiments.

\subsection{Evaluation Tasks and Datasets}
We use SentEval \cite{conneau2018senteval}, an evaluation toolkit, to evaluate the quality of sentence embeddings. Following the implementation of sentence transformers to encode the sentences, we apply dimensionality reduction methods as a post-processing step to map the embeddings into a lower-dimensional space.

Three different tasks and corresponding datasets are selected. 
\begin{description}
    \item [Semantic Textual Similarity Prediction:]
    STS-B is a set of pairwise sentences, provided with a standard benchmark of semantic similarity. 
    The standard protocol when evaluating sentence embeddings on STS datasets is to first independently encode each sentence in a pair of sentences into a sentence embedding, and then compute the similarity between the two sentences in the pair using cosine similarity.
    The predicted similarity scores are compared against the human similarity ratings in the STS datasets using some correlation coefficient.
    The Spearman rank correlation coefficient is often used for this purpose.
    It is a metric between -1 and +1, where higher positive correlations indicate better agreement with human similarity ratings.
    Therefore, a sentence embedding method that produces high positive Spearman correlations on an STS dataset is considered to be better at preserving the semantic information of the sentences in the embedding space.

    \item[Question Classification:]
    To correctly answer free-form factual questions given a large collection of texts, one must first understand the type of information that should be included in the answer to a question.
    For example, given the question \emph{What Canadian city has the largest population?}, we would classify it as having the answer type \textbf{city}, implying that only cities are valid as the candidate answers.
    \citet{li-roth-2002-learning} created a dataset from four question datasets including TREC collections and manually labelled into a hierarchical taxonomy of question types.
    Each question is classified into the the following six top-level types: ABBREVIATION, ENTITY, DESCRIPTION, HUMAN, LOCATION and NUMERIC.\footnote{Details of the full hierarchy is available at \url{https://cogcomp.seas.upenn.edu/Data/QA/QC/definition.html}.}
    The classification accuracy is measured for the effectiveness of embeddings to grasp the underlying intent of questions.

    \item[Textual Entailment:]
     Given a premise and a hypothesis expressed by two sentences, in NLI, we are expected to predict whether the hypothesis has an entailment, contradiction or a neutral relation with the premise.
    For example, given the premise ``\emph{A small girl wearing a pink jacket is riding on a carousel}'' and the hypothesis ``\emph{The carousel is moving}'', there exists an entailment relationship between the premise and the hypothesis.
    Unlike the NLP tasks discussed above, NLI prediction requires us to represent each sentence in a pair of sentences, and make a prediction for the pair as a whole instead of individual sentences in the pair.
    We use the SICK Entailment (SICK-E) dataset~\cite{marelli-etal-2014-sick} for evaluating the performance of sentence embeddings for NLI.
    SICK-E is a set of pairwise sentences, provided with a standard benchmark of relationships (contradiction, neutral and entailment). 
    The accuracy evaluates the ability of embeddings to discern semantic relations.
\end{description}

\subsection{Results}
Three factors are considered in our evaluations: tasks, sentence embeddings, and dimensions. 
Results for the SoTA simcse sentence embeddings are shown in \autoref{fig:main-results} (Other models are shown in Appendix).
Overall, PCA reports good performance with a smaller number of dimensions across tasks and sentence embeddings. 
In particular, PCA reduces dimensionality by almost $50\%$ without incurring a significant loss in task performance.
Surprisingly, when the dimensionality is greater than 150, PCA even performs better than the original sentence embeddings in STS-B. 

SVD's performance is similar to PCA in TREC, but SVD underperforms PCA for mpnet and simcse on STS-B. KPCA has comparable performance to PCA in TREC and STS-B with simcse.
However, KPCA performs poorly in TREC and SICK-E with mpnet.
The number of elements in the kernel matrix grows quadratically with the number of training instances.
In particular, we observed that the kernel matrix does not always become positive definite due to the noise in the sentence embeddings, resulting in negative eigenvalues.
Although it is possible to overcome this instability of the kernel matrix to an extent by applying small random perturbation prior to approximate eigenvalue decomposition~\cite{Halko:2010}, it still affects the performance of KPCA.

GRP never outperforms the original sentence embeddings in any task with simcse.
Its performance is identical in both inductive and transductive settings because GRP does not learn the projection from the data.
The performance of the Autoencoder with a single layer is unstable across different dimensionalities, compared to other methods, especially in TREC and SICK-E.
Overall, the transductive setting is superior to the inductive setting due to its test data awareness.

\begin{table}[t]
\centering
\resizebox{0.48\textwidth}{!}{
\begin{tabular}{l c c}
\toprule
 Method & Training Time (s) & Inference time (s)\\ \midrule
 PCA & 2.08 & \textbf{0.0049} \\ 
 KPCA & 37.98 & 0.7883 \\ 
 SVD & 2.57 & 0.0089 \\ 
 Autoencoder & 101.16 & 0.1479 \\ 
 GRP & \textbf{0.03} & 0.0080 \\ \bottomrule
\end{tabular}}
\caption{Training and inference times (wall-clock) for the different dimensionality reduction methods measured on the test set of STS-B under the inductive setting, with \textit{all-mpnet-base-v2} reduced to $300$ dimensions.}
\label{tbl:speed}
\end{table}

We see that for some tasks (e.g. simcse and mpnet on STS-B, roberta on TREC, roberta and xlm-r on SICK-E) the performance \emph{increases} when dimensions have been reduced.
Such trends have been previously observed when SVD was used to obtain lower dimensional embeddings from co-occurrence-based word embeddings~\cite{LSA,Turney_LRA,Duc:WI:2010}.
Co-occurrences between word embeddings tend to be sparse, and applying SVD collapses dimensions that are similar, thereby creating dense word embeddings that produce non-zero cosine similarity scores.
However, this does not explain the behavior observed with dense sentence embeddings used in our experiments.
Although further investigations are required as the trend is observed with specific sentence encoders and on some datasets only, we believe this is due to a form of \emph{noise reduction} due to PCA.

\autoref{tbl:speed} compares the training and inference times for the dimensionality reduction methods measured on the test data from STS-B. 
We reduce sentence embeddings produced by mpnet from 768 to 300 dimensions, selected according to \autoref{fig:main-results}, where most methods converge at.
GRP does not use training data to learn the projection, hence reports the lowest training time among all methods.

On the other hand, KPCA reports the lowest inference time.
Indeed, all PCA, SVD and GRP can be seen as multiplying a projection matrix onto the sentence embeddings corresponding to the test sentences to reduce the output dimensionality.
This operation can be naively parallelised via vectorisation, which results in extremely fast and comparable inference times for those methods.
On the other hand, Autoencoders and KPCA are both slow to train and infer with.
Autoencoders are trained with mini-batch backpropagation and require multiple iterations to converge.
Moreover, training and inference times of autoencoders further increase with the number of hidden layers.
KPCA must first compute the kernel matrix, which requires all pairwise inner products to be computed for the training instances, resulting in increased training times.

\section{Conclusion}
We evaluated unsupervised dimensionality reduction methods for pre-trained sentence embeddings using multiple NLP tasks and benchmarks under transductive and inductive settings.
The experimental results show that PCA performs consistently well across encoders and tasks.
We hope our findings will encourage the use of sentence encoders in memory/compute-constrained applications and devices.


\section{Limitations}
\label{sec:limitations}

In this paper, we considered the problem of reducing the dimensionality of sentence embeddings computed from PLMs.
All PLMs we considered were trained specifically on the English language, with the exception of \texttt{paraphrase-xlm-r-multilingual-v1 }, which is a multilingual sentence embedding model.
However, all benchmark datasets that we used for evaluations (i.e. STS-B, TREC and SICK-E) cover only English, which is a morphologically limited language.
Therefore, whether the findings reported in this paper scale to sentence embeddings for languages other than English remains an open question.
Nevertheless, we note that all dimensionality reduction methods considered in this paper are unsupervised and hence do not use any labelled data for a particular task nor a language.


Although we focused on unsupervised dimensionality reduction methods, which can be applied as a post-processing stage in this paper, as we already noted in \autoref{sec:related} there are other methods for learning models that produce lower-dimensional sentence embeddings such as supervised dimensionality reduction methods and knowledge distillation-based methods.
Conducting a comparison against all such methods is beyond the scope of this short paper and is deferred to future work.
On the other hand, our experimental results show for the first time in published literature that even with simple unsupervised dimensionality reduction methods such as PCA one can obtain surprisingly accurate lower-dimensional sentence embeddings.

\section{Ethics and Broader Impact}
\label{sec:ethics}


All datasets we used in our evaluations are collected, annotated and made publicly available in prior work on evaluating sentence embeddings.
In particular, we have not collected nor annotated any data during this project.
However, it has been reported that unfair social biases are found in STS~\cite{rudinger-etal-2017-social,webster2021measuring} and NLI~\cite{Dev:2019} datasets such as gender and racial biases.
It is possible that such biases are reflected in our evaluations.

We use a broad range of pretrained sentence embedding models as inputs in this work.
Unfortunately, it has been reported that sentence encoders have unfair social biases~\cite{may-etal-2019-measuring,kurita-etal-2019-measuring,kaneko-etal-2022-gender}.
It remains unclear how such social biases are affected by the dimensionality reduction methods we evaluate in this paper.
Although there has been prior work on evaluating social biases in text embeddings, to the best knowledge of ours, no work has evaluated the effect of dimensionality reduction in social biases.
Therefore, we consider it to be an important task to evaluate the effect on social biases due to dimensionality reduction before the methods we consider in this paper are used widely in downstream NLP tasks that require compressed sentence embeddings.

\bibliography{sentdim.bib}

\begin{thebibliography}{61}
\expandafter\ifx\csname natexlab\endcsname\relax\def\natexlab#1{#1}\fi

\bibitem[{Andrews(2016)}]{andrews2016compressing}
Martin Andrews. 2016.
\newblock Compressing word embeddings.
\newblock In \emph{International Conference on Neural Information Processing},
  pages 413--422. Springer.

\bibitem[{Anil et~al.(2018)Anil, Pereyra, Passos, Ormandi, Dahl, and
  Hinton}]{anil2018large}
Rohan Anil, Gabriel Pereyra, Alexandre Passos, Robert Ormandi, George~E. Dahl,
  and Geoffrey~E. Hinton. 2018.
\newblock Large scale distributed neural network training through online
  distillation.
\newblock In \emph{International Conference on Learning Representations}.

\bibitem[{Anowar et~al.(2021)Anowar, Sadaoui, and Selim}]{anowar2021conceptual}
Farzana Anowar, Samira Sadaoui, and Bassant Selim. 2021.
\newblock Conceptual and empirical comparison of dimensionality reduction
  algorithms (pca, kpca, lda, mds, svd, lle, isomap, le, ica, t-sne).
\newblock \emph{Computer Science Review}, 40:100378.

\bibitem[{Ayesha et~al.(2020)Ayesha, Hanif, and Talib}]{ayesha2020overview}
Shaeela Ayesha, Muhammad~Kashif Hanif, and Ramzan Talib. 2020.
\newblock Overview and comparative study of dimensionality reduction techniques
  for high dimensional data.
\newblock \emph{Information Fusion}, 59:44--58.

\bibitem[{Bingham and Mannila(2001)}]{bingham2001random}
Ella Bingham and Heikki Mannila. 2001.
\newblock Random projection in dimensionality reduction: applications to image
  and text data.
\newblock In \emph{Proceedings of the seventh ACM SIGKDD international
  conference on Knowledge discovery and data mining}, pages 245--250.

\bibitem[{Cer et~al.(2017)Cer, Diab, Agirre, Lopez-Gazpio, and
  Specia}]{cer2017semeval}
Daniel Cer, Mona Diab, Eneko Agirre, Inigo Lopez-Gazpio, and Lucia Specia.
  2017.
\newblock Semeval-2017 task 1: Semantic textual similarity-multilingual and
  cross-lingual focused evaluation.
\newblock \emph{arXiv preprint arXiv:1708.00055}.

\bibitem[{Chandar~AP et~al.(2014)Chandar~AP, Lauly, Larochelle, Khapra,
  Ravindran, Raykar, and Saha}]{chandar2014autoencoder}
Sarath Chandar~AP, Stanislas Lauly, Hugo Larochelle, Mitesh Khapra, Balaraman
  Ravindran, Vikas~C Raykar, and Amrita Saha. 2014.
\newblock An autoencoder approach to learning bilingual word representations.
\newblock \emph{Advances in neural information processing systems}, 27.

\bibitem[{Chen et~al.(2016)Chen, Mou, Xu, Li, and Jin}]{chen-EtAl:2016:P16-11}
Yunchuan Chen, Lili Mou, Yan Xu, Ge~Li, and Zhi Jin. 2016.
\newblock Compressing neural language models by sparse word representations.
\newblock In \emph{Proceedings of the 54th Annual Meeting of the Association
  for Computational Linguistics (Volume 1: Long Papers)}, pages 226--235,
  Berlin, Germany. Association for Computational Linguistics.

\bibitem[{Choi et~al.(2021)Choi, Kim, Joe, and Gwon}]{Choi:2020}
Hyunjin Choi, Judong Kim, Seongho Joe, and Youngjune Gwon. 2021.
\newblock Evaluation of bert and albert sentence embedding performance on
  downstream nlp tasks.
\newblock In \emph{2020 25th International Conference on Pattern Recognition
  (ICPR)}, pages 5482--5487.

\bibitem[{Conneau et~al.(2020)Conneau, Khandelwal, Goyal, Chaudhary, Wenzek,
  Guzman, Grave, Ott, Zettlemoyer, and Stoyanov}]{XLM-R}
Alexis Conneau, Kartikay Khandelwal, Naman Goyal, Vishrav Chaudhary, Guillaume
  Wenzek, Francisco Guzman, Edouard Grave, Myle Ott, Luke Zettlemoyer, and
  Veselin Stoyanov. 2020.
\newblock Unsupervised cross-lingual representation learning at scale.
\newblock In \emph{Proceedings of the 58th Annual Meeting of the Association
  for Computational Linguistics}, pages 8440--8451, Online. Association for
  Computational Linguistics.

\bibitem[{Conneau and Kiela(2018)}]{conneau2018senteval}
Alexis Conneau and Douwe Kiela. 2018.
\newblock Senteval: An evaluation toolkit for universal sentence
  representations.
\newblock \emph{arXiv preprint arXiv:1803.05449}.

\bibitem[{Deerwester et~al.(1990)Deerwester, Dumais, Furnas, Landauer, and
  Harshman}]{LSA}
Scott Deerwester, Susan~T. Dumais, George~W. Furnas, Thomas~K. Landauer, and
  Richard Harshman. 1990.
\newblock Indexing by latent semantic analysis.
\newblock \emph{JOURNAL OF THE AMERICAN SOCIETY FOR INFORMATION SCIENCE},
  41(6):391--407.

\bibitem[{Dev et~al.(2019)Dev, Li, Phillips, and Srikumar}]{Dev:2019}
Sunipa Dev, Tao Li, Jeff Phillips, and Vivek Srikumar. 2019.
\newblock {O}n {M}easuring and {M}itigating {B}iased {I}nferences of {W}ord
  {E}mbeddings.

\bibitem[{Devlin et~al.(2018)Devlin, Chang, Lee, and
  Toutanova}]{devlin2018bert}
Jacob Devlin, Ming-Wei Chang, Kenton Lee, and Kristina Toutanova. 2018.
\newblock Bert: Pre-training of deep bidirectional transformers for language
  understanding.
\newblock \emph{arXiv preprint arXiv:1810.04805}.

\bibitem[{Duc et~al.(2010)Duc, Bollegala, and Ishizuka}]{Duc:WI:2010}
Nguyen~Tuan Duc, Danushka Bollegala, and Mitsuru Ishizuka. 2010.
\newblock Using relational similarity between word pairs for latent relational
  search on the web.
\newblock In \emph{IEEE/WIC/ACM International Conference on Web Intelligence
  and Intelligent Agent Technology}, pages 196 -- 199.

\bibitem[{Dzikovska et~al.(2013)Dzikovska, Nielsen, Brew, Leacock, Giampiccolo,
  Bentivogli, Clark, Dagan, and Dang}]{dzikovska-etal-2013-semeval}
Myroslava Dzikovska, Rodney Nielsen, Chris Brew, Claudia Leacock, Danilo
  Giampiccolo, Luisa Bentivogli, Peter Clark, Ido Dagan, and Hoa~Trang Dang.
  2013.
\newblock {S}em{E}val-2013 task 7: The joint student response analysis and 8th
  recognizing textual entailment challenge.
\newblock In \emph{Second Joint Conference on Lexical and Computational
  Semantics (*{SEM}), Volume 2: Proceedings of the Seventh International
  Workshop on Semantic Evaluation ({S}em{E}val 2013)}, pages 263--274, Atlanta,
  Georgia, USA. Association for Computational Linguistics.

\bibitem[{Gao et~al.(2021{\natexlab{a}})Gao, Yao, and
  Chen}]{gao-etal-2021-simcse}
Tianyu Gao, Xingcheng Yao, and Danqi Chen. 2021{\natexlab{a}}.
\newblock {S}im{CSE}: Simple contrastive learning of sentence embeddings.
\newblock In \emph{Proceedings of the 2021 Conference on Empirical Methods in
  Natural Language Processing}, pages 6894--6910, Online and Punta Cana,
  Dominican Republic. Association for Computational Linguistics.

\bibitem[{Gao et~al.(2021{\natexlab{b}})Gao, Yao, and Chen}]{gao2021simcse}
Tianyu Gao, Xingcheng Yao, and Danqi Chen. 2021{\natexlab{b}}.
\newblock Simcse: Simple contrastive learning of sentence embeddings.
\newblock \emph{arXiv preprint arXiv:2104.08821}.

\bibitem[{Gupta et~al.(2019)Gupta, Giesselbach, R{\"u}ping, and
  Bauckhage}]{gupta-etal-2019-improving}
Vishwani Gupta, Sven Giesselbach, Stefan R{\"u}ping, and Christian Bauckhage.
  2019.
\newblock \href {https://doi.org/10.18653/v1/W19-4323} {Improving word
  embeddings using kernel {PCA}}.
\newblock In \emph{Proceedings of the 4th Workshop on Representation Learning
  for NLP (RepL4NLP-2019)}, pages 200--208, Florence, Italy. Association for
  Computational Linguistics.

\bibitem[{Halko et~al.(2010)Halko, Martinsson, and Tropp}]{Halko:2010}
N.~Halko, P.~G. Martinsson, and J.~A. Tropp. 2010.
\newblock Finding structure with randomness: Probabilistic algorithms for
  constructung approximate matrix decompositions.
\newblock \emph{SIAM REVIEW}, 53(2):217 -- 288.

\bibitem[{Han et~al.(2016)Han, Mao, and Dally}]{Han:2016}
Song Han, Huizi Mao, and William~J. Dally. 2016.
\newblock Deep compression: Compressing deep neural network with pruning,
  trained quantization and huffman coding.
\newblock In \emph{4th International Conference on Learning Representations,
  {ICLR} 2016, San Juan, Puerto Rico, May 2-4, 2016, Conference Track
  Proceedings}.

\bibitem[{Han et~al.(2015)Han, Pool, Tran, and Dally}]{Han:2015}
Song Han, Jeff Pool, John Tran, and William~J. Dally. 2015.
\newblock Learning both weights and connections for efficient neural networks.
\newblock In \emph{Proceedings of the 28th International Conference on Neural
  Information Processing Systems - Volume 1}, NIPS'15, pages 1135--1143,
  Cambridge, MA, USA. MIT Press.

\bibitem[{Hao et~al.(2019)Hao, Liu, Wu, and Lv}]{Hao:2019}
Yu~Hao, Xien Liu, Ji~Wu, and Ping Lv. 2019.
\newblock Exploiting sentence embedding for medical question answering.
\newblock In \emph{Proceedings of the Thirty-Third AAAI Conference on
  Artificial Intelligence and Thirty-First Innovative Applications of
  Artificial Intelligence Conference and Ninth AAAI Symposium on Educational
  Advances in Artificial Intelligence}, AAAI'19/IAAI'19/EAAI'19. AAAI Press.

\bibitem[{Hovy et~al.(2001)Hovy, Gerber, Hermjakob, Lin, and
  Ravichandran}]{hovy-etal-2001-toward}
Eduard Hovy, Laurie Gerber, Ulf Hermjakob, Chin-Yew Lin, and Deepak
  Ravichandran. 2001.
\newblock \href {https://www.aclweb.org/anthology/H01-1069} {Toward
  semantics-based answer pinpointing}.
\newblock In \emph{Proceedings of the First International Conference on Human
  Language Technology Research}.

\bibitem[{Jiao et~al.(2020)Jiao, Yin, Shang, Jiang, Chen, Li, Wang, and
  Liu}]{jiao-etal-2020-tinybert}
Xiaoqi Jiao, Yichun Yin, Lifeng Shang, Xin Jiang, Xiao Chen, Linlin Li, Fang
  Wang, and Qun Liu. 2020.
\newblock {T}iny{BERT}: Distilling {BERT} for natural language understanding.
\newblock In \emph{Findings of the Association for Computational Linguistics:
  EMNLP 2020}, pages 4163--4174, Online. Association for Computational
  Linguistics.

\bibitem[{Kaneko et~al.(2022)Kaneko, Imankulova, Bollegala, and
  Okazaki}]{kaneko-etal-2022-gender}
Masahiro Kaneko, Aizhan Imankulova, Danushka Bollegala, and Naoaki Okazaki.
  2022.
\newblock Gender bias in masked language models for multiple languages.
\newblock In \emph{Proceedings of the 2022 Conference of the North American
  Chapter of the Association for Computational Linguistics: Human Language
  Technologies}, pages 2740--2750, Seattle, United States. Association for
  Computational Linguistics.

\bibitem[{Kim et~al.(2020)Kim, Kim, and Lee}]{kim-etal-2020-adaptive}
Yeachan Kim, Kang-Min Kim, and SangKeun Lee. 2020.
\newblock \href {https://doi.org/10.18653/v1/2020.acl-main.364} {Adaptive
  compression of word embeddings}.
\newblock In \emph{Proceedings of the 58th Annual Meeting of the Association
  for Computational Linguistics}, pages 3950--3959, Online. Association for
  Computational Linguistics.

\bibitem[{Kong et~al.(2022)Kong, Khadanga, Li, Gupta, Zhang, Xu, and
  Bendersky}]{Weize:2022}
Weize Kong, Swaraj Khadanga, Cheng Li, Shaleen~Kumar Gupta, Mingyang Zhang,
  Wensong Xu, and Michael Bendersky. 2022.
\newblock \href {https://doi.org/10.1145/3534678.3539137} {Multi-aspect dense
  retrieval}.
\newblock In \emph{Proceedings of the 28th ACM SIGKDD Conference on Knowledge
  Discovery and Data Mining}, KDD '22, page 3178–3186, New York, NY, USA.
  Association for Computing Machinery.

\bibitem[{Kurita et~al.(2019)Kurita, Vyas, Pareek, Black, and
  Tsvetkov}]{kurita-etal-2019-measuring}
Keita Kurita, Nidhi Vyas, Ayush Pareek, Alan~W Black, and Yulia Tsvetkov. 2019.
\newblock Measuring bias in contextualized word representations.
\newblock In \emph{Proceedings of the First Workshop on Gender Bias in Natural
  Language Processing}, pages 166--172, Florence, Italy. Association for
  Computational Linguistics.

\bibitem[{Lee et~al.(2019)Lee, Ajanthan, and Torr}]{Lee:2019}
Namhoon Lee, Thalaiyasingam Ajanthan, and Philip H.~S. Torr. 2019.
\newblock Snip: single-shot network pruning based on connection sensitivity.
\newblock In \emph{7th International Conference on Learning Representations,
  {ICLR} 2019, New Orleans, LA, USA, May 6-9, 2019}. OpenReview.net.

\bibitem[{Levy et~al.(2015)Levy, Goldberg, and Dagan}]{levy2015improving}
Omer Levy, Yoav Goldberg, and Ido Dagan. 2015.
\newblock Improving distributional similarity with lessons learned from word
  embeddings.
\newblock \emph{Transactions of the association for computational linguistics},
  3:211--225.

\bibitem[{Li et~al.(2017)Li, Kadav, Durdanovic, Samet, and Graf}]{Li:2017}
Hao Li, Asim Kadav, Igor Durdanovic, Hanan Samet, and Hans~Peter Graf. 2017.
\newblock Pruning filters for efficient convnets.
\newblock In \emph{International Conference on Learning Representations}.

\bibitem[{Li and Roth(2002)}]{li-roth-2002-learning}
Xin Li and Dan Roth. 2002.
\newblock \href {https://www.aclweb.org/anthology/C02-1150} {Learning question
  classifiers}.
\newblock In \emph{{COLING} 2002: The 19th International Conference on
  Computational Linguistics}.

\bibitem[{Liu et~al.(2019)Liu, Ott, Goyal, Du, Joshi, Chen, Levy, Lewis,
  Zettlemoyer, and Stoyanov}]{liu2019roberta}
Yinhan Liu, Myle Ott, Naman Goyal, Jingfei Du, Mandar Joshi, Danqi Chen, Omer
  Levy, Mike Lewis, Luke Zettlemoyer, and Veselin Stoyanov. 2019.
\newblock Roberta: A robustly optimized bert pretraining approach.
\newblock \emph{arXiv preprint arXiv:1907.11692}.

\bibitem[{Marelli et~al.(2014)Marelli, Menini, Baroni, Bentivogli, Bernardi,
  and Zamparelli}]{marelli-etal-2014-sick}
Marco Marelli, Stefano Menini, Marco Baroni, Luisa Bentivogli, Raffaella
  Bernardi, and Roberto Zamparelli. 2014.
\newblock A {SICK} cure for the evaluation of compositional distributional
  semantic models.
\newblock In \emph{Proceedings of the Ninth International Conference on
  Language Resources and Evaluation ({LREC}'14)}, pages 216--223, Reykjavik,
  Iceland. European Language Resources Association (ELRA).

\bibitem[{May et~al.(2019)May, Wang, Bordia, Bowman, and
  Rudinger}]{may-etal-2019-measuring}
Chandler May, Alex Wang, Shikha Bordia, Samuel~R. Bowman, and Rachel Rudinger.
  2019.
\newblock On measuring social biases in sentence encoders.
\newblock In \emph{Proceedings of the 2019 Conference of the North {A}merican
  Chapter of the Association for Computational Linguistics: Human Language
  Technologies, Volume 1 (Long and Short Papers)}, pages 622--628, Minneapolis,
  Minnesota. Association for Computational Linguistics.

\bibitem[{Menon et~al.(2022)Menon, Jayasumana, Rawat, Kim, Reddi, and
  Kumar}]{pmlr-v162-menon22a}
Aditya Menon, Sadeep Jayasumana, Ankit~Singh Rawat, Seungyeon Kim, Sashank
  Reddi, and Sanjiv Kumar. 2022.
\newblock In defense of dual-encoders for neural ranking.
\newblock In \emph{Proceedings of the 39th International Conference on Machine
  Learning}, volume 162 of \emph{Proceedings of Machine Learning Research},
  pages 15376--15400. PMLR.

\bibitem[{Mikolov et~al.(2013)Mikolov, Chen, and Dean}]{Milkov:2013}
Tomas Mikolov, Kai Chen, and Jeffrey Dean. 2013.
\newblock Efficient estimation of word representation in vector space.
\newblock In \emph{Proc. of International Conference on Learning
  Representations}.

\bibitem[{Nguyen et~al.(2016)Nguyen, Rosenberg, Song, Gao, Tiwary, Majumder,
  and Deng}]{nguyen2016ms}
Tri Nguyen, Mir Rosenberg, Xia Song, Jianfeng Gao, Saurabh Tiwary, Rangan
  Majumder, and Li~Deng. 2016.
\newblock Ms marco: A human generated machine reading comprehension dataset.
\newblock In \emph{CoCo@ NIPs}.

\bibitem[{Nigam et~al.(2019)Nigam, Song, Mohan, Lakshman, Weitian, Ding,
  Shingavi, Teo, Gu, and Yin}]{SPS}
Priyanka Nigam, Yiwei Song, Vijai Mohan, Vihan Lakshman, Weitian, Ding, Ankit
  Shingavi, Choon~Hui Teo, Hao Gu, and Bing Yin. 2019.
\newblock {S}emantic {P}roduct {S}earch.

\bibitem[{Palangi et~al.(2016)Palangi, Deng, Shen, Gao, He, Chen, Song, and
  Ward}]{Palangi:2016}
Hamid Palangi, Li~Deng, Yelong Shen, Jianfeng Gao, Xiaodong He, Jianshu Chen,
  Xinying Song, and Rabab Ward. 2016.
\newblock \href {https://doi.org/10.1109/TASLP.2016.2520371} {Deep sentence
  embedding using long short-term memory networks: Analysis and application to
  information retrieval}.
\newblock \emph{IEEE/ACM Trans. Audio, Speech and Lang. Proc.},
  24(4):694–707.

\bibitem[{Pennington et~al.(2014)Pennington, Socher, and Manning}]{Glove}
Jeffery Pennington, Richard Socher, and Christopher~D. Manning. 2014.
\newblock Glove: global vectors for word representation.
\newblock In \emph{Proc. of EMNLP}, pages 1532--1543.

\bibitem[{Petersen and Petersen(2012)}]{MatrixCookbook}
Kaare~Brandt Petersen and Michael~Syskind Petersen. 2012.
\newblock \emph{The Matrix Cookbook}.
\newblock Online.

\bibitem[{Raunak et~al.(2019)Raunak, Gupta, and Metze}]{raunak2019effective}
Vikas Raunak, Vivek Gupta, and Florian Metze. 2019.
\newblock Effective dimensionality reduction for word embeddings.
\newblock In \emph{Proceedings of the 4th Workshop on Representation Learning
  for NLP (RepL4NLP-2019)}, pages 235--243.

\bibitem[{Reddy et~al.(2020)Reddy, Reddy, Lakshmanna, Kaluri, Rajput,
  Srivastava, and Baker}]{reddy2020analysis}
G~Thippa Reddy, M~Praveen~Kumar Reddy, Kuruva Lakshmanna, Rajesh Kaluri,
  Dharmendra~Singh Rajput, Gautam Srivastava, and Thar Baker. 2020.
\newblock Analysis of dimensionality reduction techniques on big data.
\newblock \emph{IEEE Access}, 8:54776--54788.

\bibitem[{Reimers and
  Gurevych(2019{\natexlab{a}})}]{reimers-gurevych-2019-sentence}
Nils Reimers and Iryna Gurevych. 2019{\natexlab{a}}.
\newblock Sentence-{BERT}: Sentence embeddings using {S}iamese {BERT}-networks.
\newblock In \emph{Proceedings of the 2019 Conference on Empirical Methods in
  Natural Language Processing and the 9th International Joint Conference on
  Natural Language Processing (EMNLP-IJCNLP)}, pages 3980--3990, Hong Kong,
  China. Association for Computational Linguistics.

\bibitem[{Reimers and
  Gurevych(2019{\natexlab{b}})}]{reimers-2019-sentence-bert}
Nils Reimers and Iryna Gurevych. 2019{\natexlab{b}}.
\newblock \href {http://arxiv.org/abs/1908.10084} {Sentence-bert: Sentence
  embeddings using siamese bert-networks}.
\newblock In \emph{Proceedings of the 2019 Conference on Empirical Methods in
  Natural Language Processing}. Association for Computational Linguistics.

\bibitem[{Rudinger et~al.(2017)Rudinger, May, and
  Van~Durme}]{rudinger-etal-2017-social}
Rachel Rudinger, Chandler May, and Benjamin Van~Durme. 2017.
\newblock Social bias in elicited natural language inferences.
\newblock In \emph{Proceedings of the First {ACL} Workshop on Ethics in Natural
  Language Processing}, pages 74--79, Valencia, Spain. Association for
  Computational Linguistics.

\bibitem[{Sanh et~al.(2019)Sanh, Debut, Chaumond, and Wolf}]{DistilBERT}
Victor Sanh, Lysandre Debut, Julien Chaumond, and Thomas Wolf. 2019.
\newblock Distilbert, a distilled version of bert: smaller, faster, cheaper and
  lighter.
\newblock In \emph{Proc. of the 5th Workshop on Energy Efficient Machine
  Learning and Cognitive Computing at NeurIPS-2019}.

\bibitem[{Schmidt(2018)}]{schmidt2018stable}
Benjamin Schmidt. 2018.
\newblock Stable random projection: Lightweight, general-purpose dimensionality
  reduction for digitized libraries.
\newblock \emph{Journal of Cultural Analytics}, 3(1):11033.

\bibitem[{Sch{\"o}lkopf et~al.(1998)Sch{\"o}lkopf, Smola, and
  M{\"u}ller}]{kernelPCA}
Bernhard Sch{\"o}lkopf, Alexander Smola, and Klaus-Robert M{\"u}ller. 1998.
\newblock {Nonlinear Component Analysis as a Kernel Eigenvalue Problem}.
\newblock \emph{Neural Computation}, 10(5):1299--1319.

\bibitem[{Shu and Nakayama(2018)}]{shu2018compressing}
Raphael Shu and Hideki Nakayama. 2018.
\newblock \href {https://openreview.net/forum?id=BJRZzFlRb} {Compressing word
  embeddings via deep compositional code learning}.
\newblock In \emph{International Conference on Learning Representations}.

\bibitem[{Socher et~al.(2011)Socher, Huang, Pennin, Manning, and
  Ng}]{socher2011dynamic}
Richard Socher, Eric Huang, Jeffrey Pennin, Christopher~D Manning, and Andrew
  Ng. 2011.
\newblock Dynamic pooling and unfolding recursive autoencoders for paraphrase
  detection.
\newblock \emph{Advances in neural information processing systems}, 24.

\bibitem[{Song et~al.(2020)Song, Tan, Qin, Lu, and Liu}]{2004.09297}
Kaitao Song, Xu~Tan, Tao Qin, Jianfeng Lu, and Tie-Yan Liu. 2020.
\newblock {M}{P}{N}et: {M}asked and {P}ermuted {P}re-training for {L}anguage
  {U}nderstanding.

\bibitem[{Turney(2005)}]{Turney_LRA}
P.D. Turney. 2005.
\newblock Measuring semantic similarity by latent relational analysis.
\newblock In \emph{Proc. of IJCAI'05}, pages 1136--1141.

\bibitem[{Vincent et~al.(2008)Vincent, Larochelle, Bengio, and
  Manzagol}]{Vincent:ICML:2008}
Pascal Vincent, Hugo Larochelle, Yoshua Bengio, and Pierre-Antonie Manzagol.
  2008.
\newblock Extracting and composing robust features with denoising autoencoders.
\newblock In \emph{ICML'08}, pages 1096 -- 1103.

\bibitem[{Wang et~al.(2017)Wang, Finch, Utiyama, and
  Sumita}]{wang-etal-2017-sentence}
Rui Wang, Andrew Finch, Masao Utiyama, and Eiichiro Sumita. 2017.
\newblock Sentence embedding for neural machine translation domain adaptation.
\newblock In \emph{Proceedings of the 55th Annual Meeting of the Association
  for Computational Linguistics (Volume 2: Short Papers)}, pages 560--566,
  Vancouver, Canada. Association for Computational Linguistics.

\bibitem[{Wang et~al.(2016)Wang, Yao, and Zhao}]{wang2016auto}
Yasi Wang, Hongxun Yao, and Sicheng Zhao. 2016.
\newblock Auto-encoder based dimensionality reduction.
\newblock \emph{Neurocomputing}, 184:232--242.

\bibitem[{Webster et~al.(2021)Webster, Wang, Tenney, Beutel, Pitler, Pavlick,
  Chen, Chi, and Petrov}]{webster2021measuring}
Kellie Webster, Xuezhi Wang, Ian Tenney, Alex Beutel, Emily Pitler, Ellie
  Pavlick, Jilin Chen, Ed~Chi, and Slav Petrov. 2021.
\newblock Measuring and reducing gendered correlations in pre-trained models.

\bibitem[{Yokote et~al.(2012)Yokote, Bollegala, and
  Ishizuka}]{Yokote:AAAI:2012}
Ken-ichi Yokote, Danushka Bollegala, and Mitsuru Ishizuka. 2012.
\newblock Similarity is not entailment -- jointly learning similarity
  transformations for textual entailment.
\newblock In \emph{Proc. of the National Conference on Artificial Intelligence
  (AAAI)}, pages 1720 -- 1726.

\bibitem[{Zhao et~al.(2022)Zhao, Yu, Wu, and Li}]{zhao2022compressing}
Xuandong Zhao, Zhiguo Yu, Ming Wu, and Lei Li. 2022.
\newblock Compressing sentence representation for semantic retrieval via
  homomorphic projective distillation.
\newblock \emph{arXiv preprint arXiv:2203.07687}.

\end{thebibliography}
\bibliographystyle{lrec-coling2024-natbib}

\appendix
\section*{Appendix}

\section{Dimensionality Reduction Methods}
\label{sec:sup:dim-reduct}

\subsection{Truncated SVD}
Singular value decomposition (SVD) is a matrix factorization method that finds a lower-rank approximation to the original matrix by minimizing the squared Frobenious norm between the original matrix and its low-rank factorization.
SVD generalizes the eigenvalue decomposition of square matrices to rectangular matrices.
Specifically, given a matrix $\mat{X} (\in R^{m \times n})$ where $m$ is the number of observations and $n$ is the number of features, SVD decomposes it into the product of three matrices as given by \eqref{eq:SVD}.
\begin{align}
    \label{eq:SVD}
    \mat{X} = \mat{U}{\Sigma}\mat{V}^{*}
\end{align}
Here,  $\mat{U}$ is an $m \times m$ unitary matrix (columns are known as left singular vectors), $\mathbf{\Sigma}$ is an $m \times n$ diagonal matrix (diagonal values are known as singular values), $\mat{V}$ is an $n \times n$ unitary matrix, and $\mat{V}^*$ is the conjugate transpose of $\mat{V}$ (rows are known as right singular vectors). 
To obtain a lower-dimensional approximation, suppose the projection dimension is selected as $k$ ($k \leq n$) and the first $k$ columns of $\mat{V}^*$ of the training vector space are retained as a $n \times k$ matrix $\mat{V}_k$, known as the projection matrix. The target vector space of the original $m \times n$ matrix $\mat{X}$ will be projected as $\mat{X}\mat{V}_k$ The final dimensionality then becomes $m \times k$. For each sample, dimensionality is reduced to $k$.

\subsection{PCA}
Principal component analysis (PCA) is a process of projecting data into a new basis, by computing the principal components. PCA is an unsupervised orthogonal statistical technique where the interactions between features can be extracted by computing the covariance matrix of the original data. Specifically, suppose an $m \times n$ matrix $\mathbf{X}$ is represented by the original points where $m$ is the number of observations and $n$ is the number of features. PCA first standardizes $\mathbf{X}$. Suppose the target projection dimension is $k$ ($k \leq n$). The normalised largest $k$ eigenvectors of the covariance matrix ($\mathbf{X}\mathbf{X}^\mathrm{T}$) are chosen to form the new axes, forming a $n \times k$ vector space $\mathbf{U}$. The corresponding eigenvalues are used to order the eigenvectors. The variance of the projected data in the chosen direction is maximised to preserve the diversity of data in each dimension. This is, the information structure in the data is retained to the maximum extent. The target vector space of the original $m \times n$ matrix $\mathbf{X}$ will be projected as 
\begin{align}
   \mathbf{X}\mathbf{U} 
\end{align}
The final dimensionality then becomes $m \times k$.

\subsection{Kernel PCA}
Kernel principal component analysis (KPCA) is a non-linear dimensionality reduction method, which is an extension of PCA. Since PCA is limited to linear projections, KPCA uses kernel functions to deal with non-linear data and make it linearly separable \cite{anowar2021conceptual}. KPCA works by first projecting original data to higher-dimensional space with kernel functions such as Polynomial and Sigmoid kernels. Specifically, suppose an $m \times n$ matrix $\mathbf{X}$ is represented by the original points. Suppose the target projection dimension is $k$ ($k \leq n$). KPCA maps $\mathbf{X}$ into a higher feature space by function $\Phi: \mathbb{R}^n \rightarrow \mathbb{R}^d$ ($d>n$), where $\Phi$ makes data linearly separable. Then a kernel matrix is generated $$\mathbf{K}=\Phi(\mathbf{X})^T \Phi(\mathbf{X}).$$ After centering the kernel matrix, eigendecomposition is utilized to compute the eigenvectors and eigenvalues like PCA. Procedures of PCA is thus applied in the following calculation to reduce the sample dimension to $k$ ($k<n$).

\subsection{Gaussian random projection}
Gaussian Random Projection is a technique used to reduce the dimensions by projecting the original high-dimensional input space using a randomly generated matrix. The core idea behind random projection is Johnson-Lindenstrauss lemma, which states that distances between points can be nearly preserved when embedding high-dimensional space into a much lower-dimensional space. Specifically, suppose the original dataset is defined as a $m \times n$ matrix $\mathbf{X}$. Suppose the target projection dimension is $k$ ($k \leq n$). To obtain a lower-dimensional projection of the input data, first a set of $k$-dimensional normally distributed random vectors is generated. Then the projection matrix $n \times k$ matrix $\mathbf{R}$ is produced by stacking those vectors. 
Finally, the original matrix $m \times n$ matrix $\mathbf{X}$ will be projected as in \eqref{eq:gaussian}.
\begin{align}
    \label{eq:gaussian}
    \mathbf{X}\mathbf{R}
\end{align}
The final dimensionality then becomes $m \times k$.

\subsection{Autoencoders}
Autoencoders can be seen as a non-linear version of PCA, where an encoder network first projects the inputs to a (possibly lower-dimensional) space.
Next, some non-linear function is applied elementwise on the encoded input.
Finally, a decoder network attempts to reconstruct the input.
The encoder and decoder network parameters are jointly learned such that the reconstruction loss (e.g. measured by the squared $\ell_2$ distance between the input and its corresponding reconstruction) is minimized.
 To compress the input, we can constrain the dimensionality of the hidden layer~\cite{wang2016auto} to be smaller than the input dimensionality.
 
 \section{Sentence Encoders}
 \label{sec:encoders}
 

SentenceTransformers \cite{reimers-2019-sentence-bert} is a Python framework to embed sentences with state-of-the-art models. Spcific sentence encoders we use are \textit{all-mpnet-base-v2} \cite{2004.09297}, \textit{msmarco-roberta-base-v2} \cite{reimers-2019-sentence-bert}, \textit{paraphrase-xlm-r-multilingual-v1} \cite{XLM-R}, \textit{stsb-bert-base} \cite{reimers-2019-sentence-bert} and \textit{stsb-bert-large} \cite{reimers-2019-sentence-bert}. Additionally, \textit{all-mpnet-base-v2} is fine-tuned on a sentence pairs dataset, based on \textit{microsoft/mpnet-base} model \cite{2004.09297}. \textit{stsb-bert-base} and \textit{stsb-bert-large} are fine-tuned on STS-B dataset, based on BERT \cite{devlin2018bert}. \textit{msmarco-roberta-base-v2} is fine-tuned on Microsoft Machine Reading Comprehension (MS MARCO) dataset \cite{nguyen2016ms}, based on RoBERTa \cite{liu2019roberta}. Princeton-NLP provides a Python-version sentence encoder \textit{sup-simcse-roberta-large}~\cite{gao2021simcse}.


 
 


 \section{Experiment Results for Sentence Encoders}
\label{sec:appendix:experiments}
 
 \begin{figure*}[t!]
    \vspace{-0.5cm}
    \centering
    \subfigure[Transductive setting]{
        \rotatebox{90}{\footnotesize{~~~~~~~~~~~sup-simcse-roberta-large~~~~~~~~~~~~~~~~~~~~~~~~~~~all-mpnet-base-v2~~~~~~~~~~~~~~~~~~~~~~~~~~~~~~~~~~~stsb-bert-large~~~~~~~~~~~~~~~~~~~~~~~~~~~~~~~~~~~~~~~stsb-bert-base}}
        \begin{minipage}[b]{0.42\linewidth}
        \includegraphics[width=1.07\linewidth]{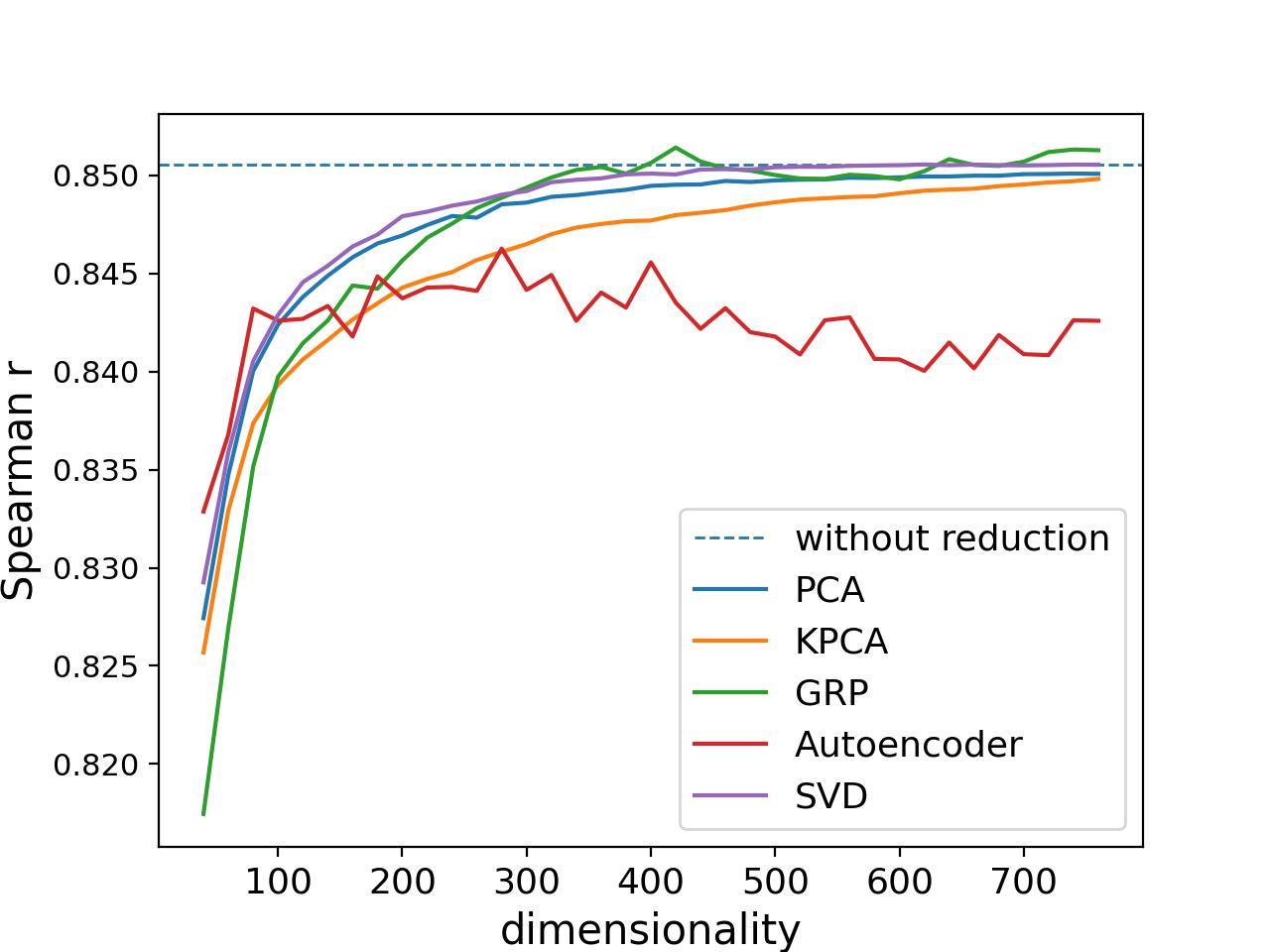}
        \vspace{0cm}
        \includegraphics[width=1.07\linewidth]{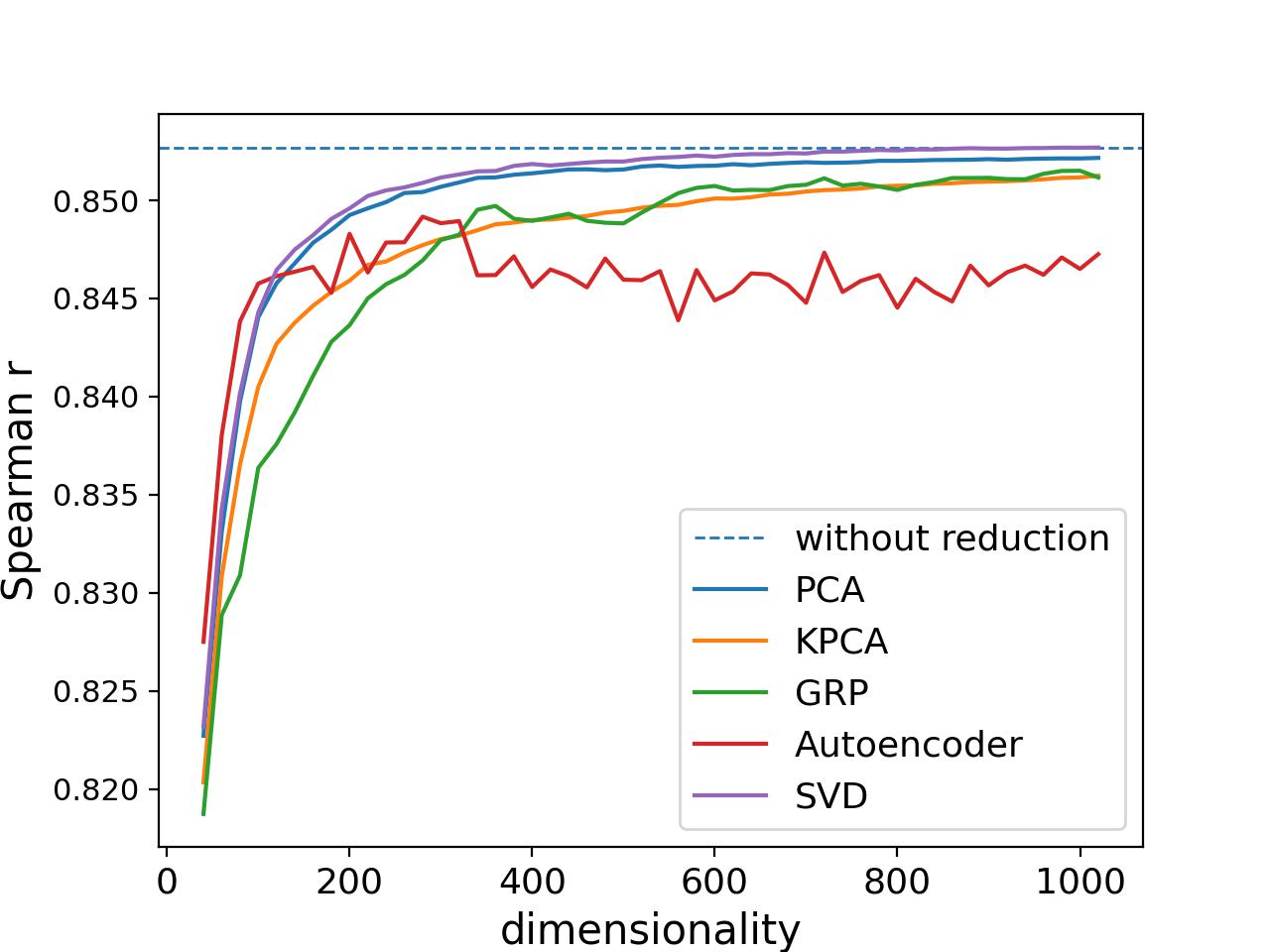}
        \vspace{0cm}
        \includegraphics[width=1.07\linewidth]{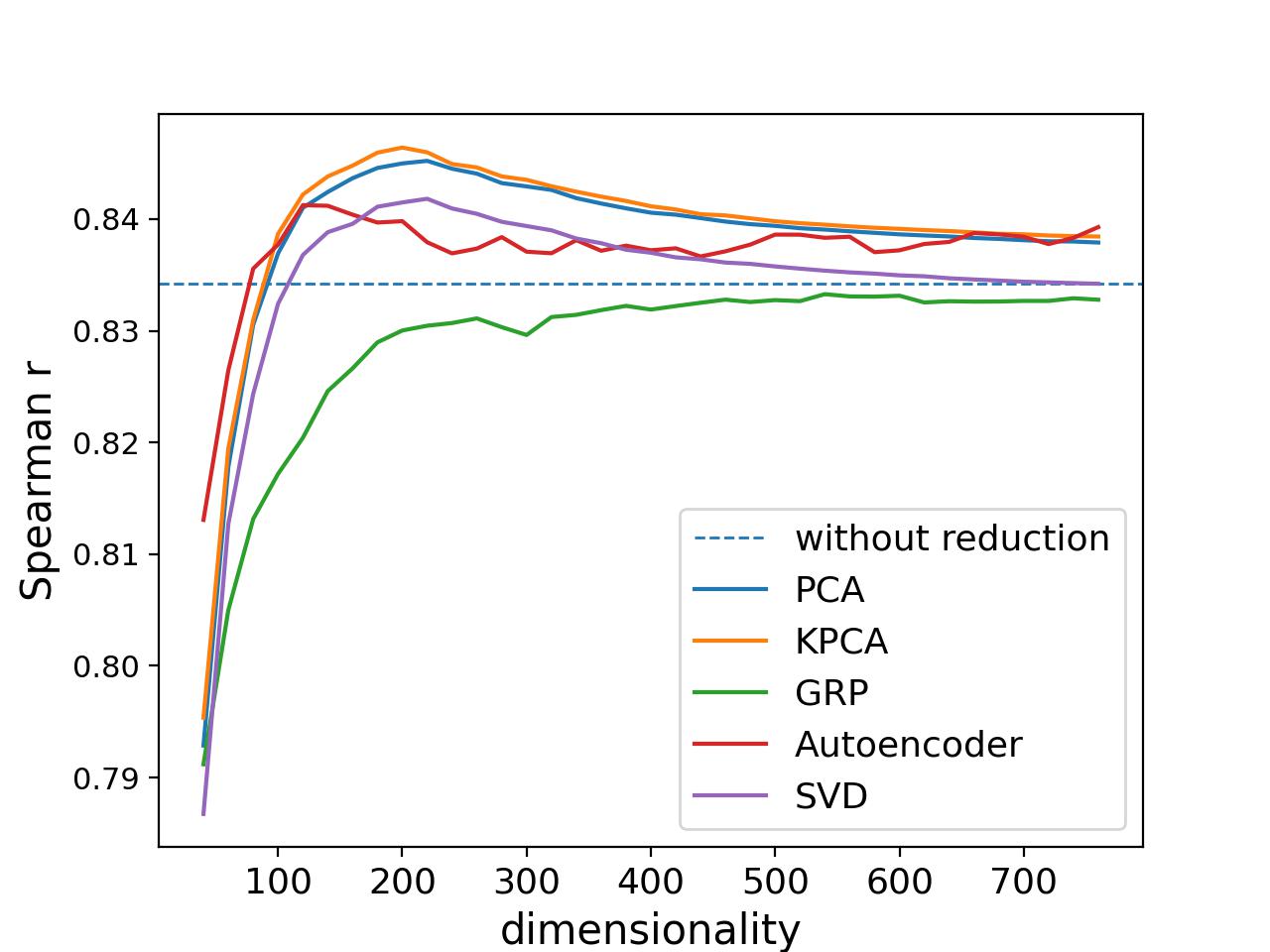}
        \vspace{0cm}
        \includegraphics[width=1.07\linewidth]{picture//trans_STSB_SimCSE.jpg}
        \end{minipage}
    }
    \subfigure[Inductive setting]{
        \begin{minipage}[b]{0.42\linewidth} 
        \includegraphics[width=1.07\linewidth]{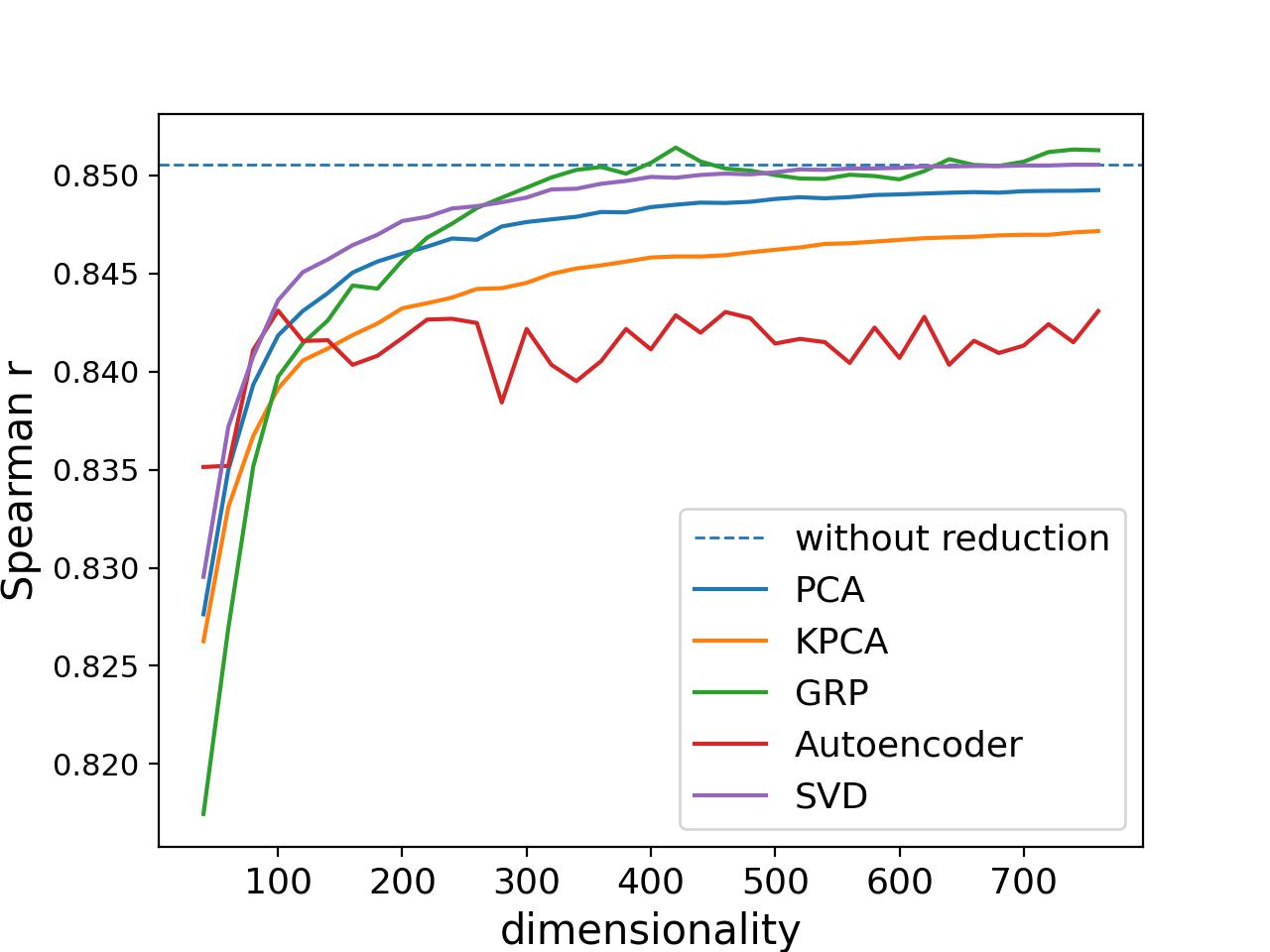}
        \vspace{0cm}
        \includegraphics[width=1.07\linewidth]{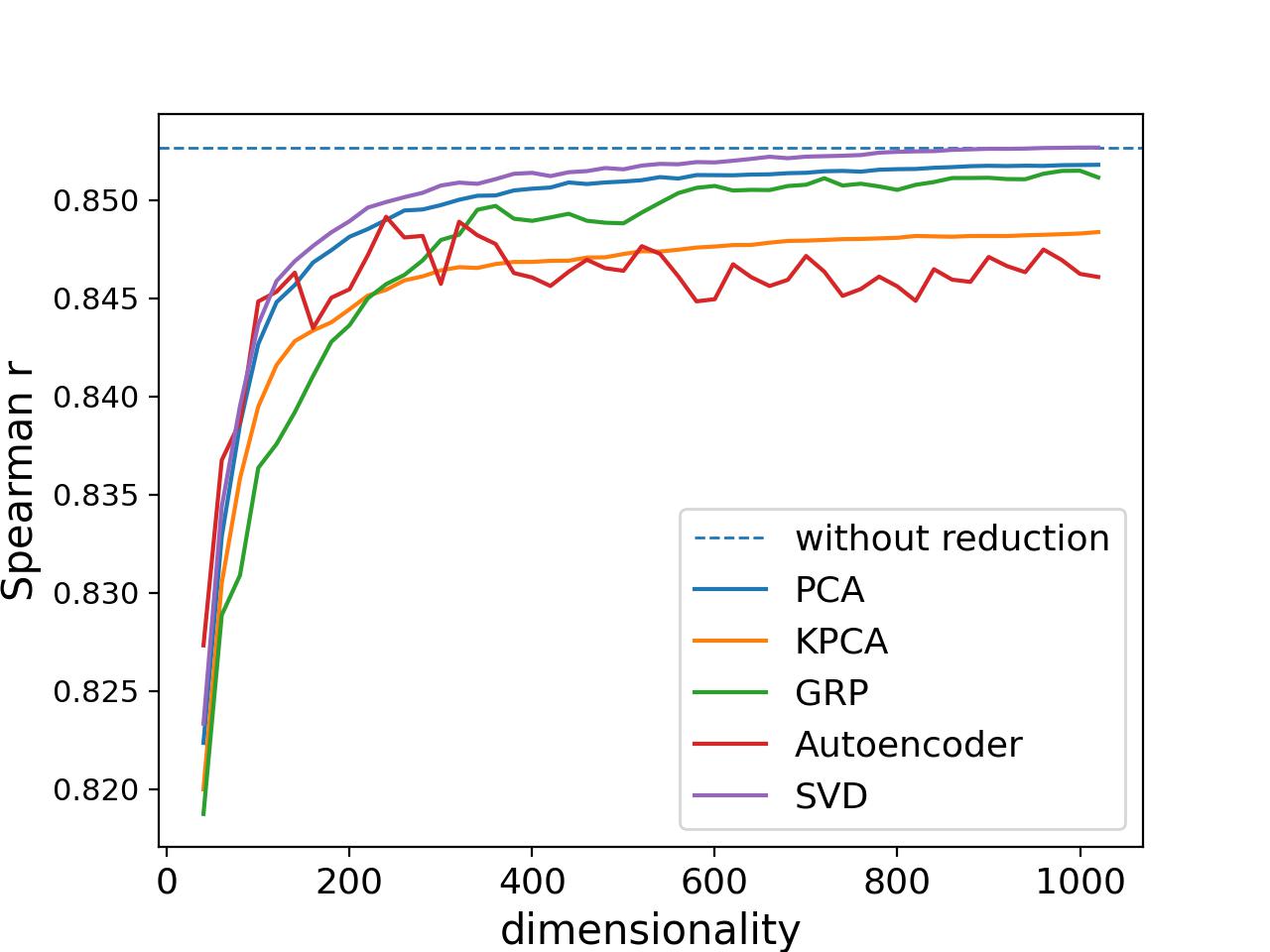}
        \vspace{0cm}
        \includegraphics[width=1.07\linewidth]{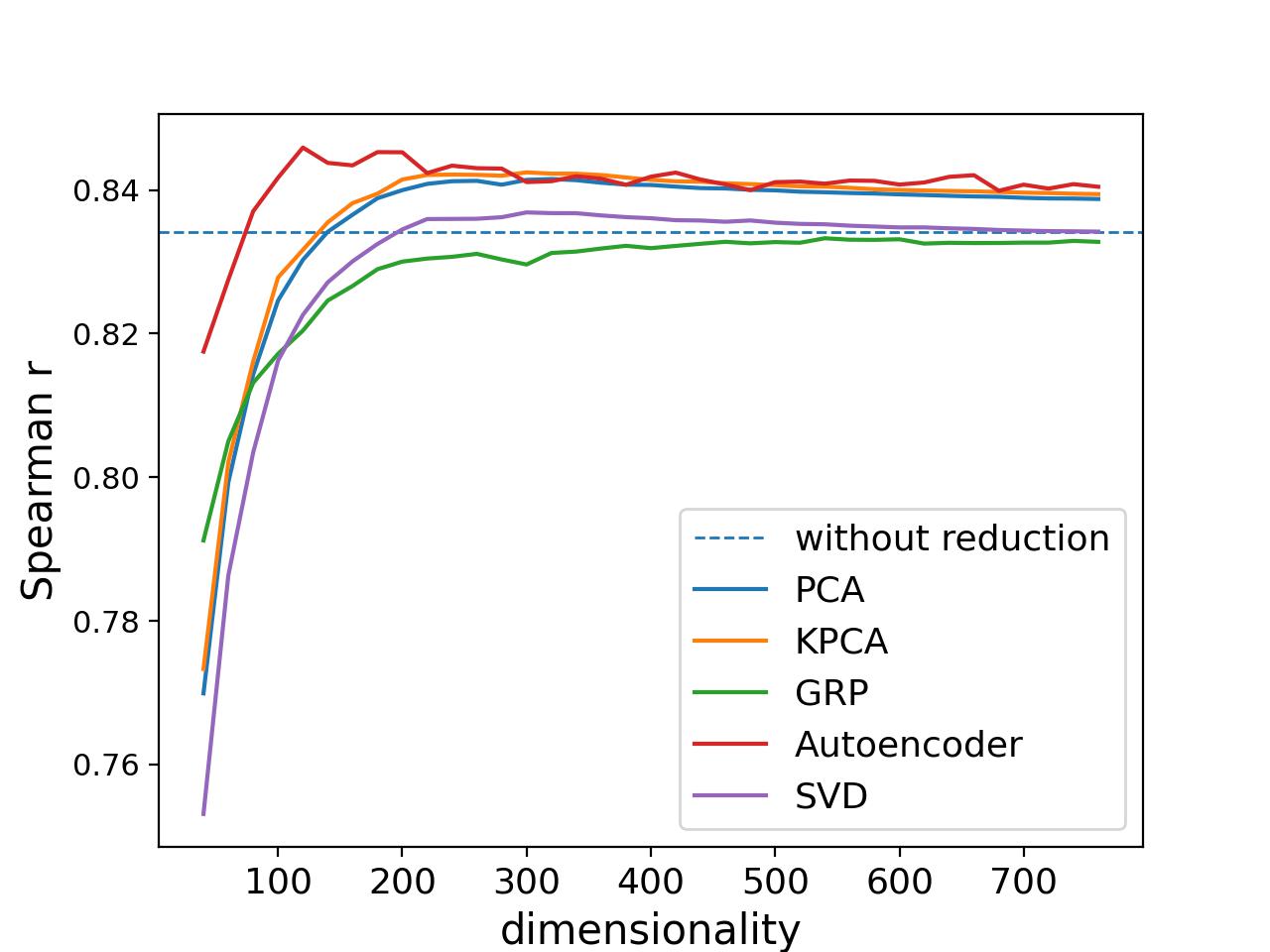}
        \vspace{0cm}
        \includegraphics[width=1.07\linewidth]{picture//in_STSB_SimCSE.jpg}
        \end{minipage}
    }
    \caption{Spearman correlation coefficients on STS-B vs. the dimensionality of the sentence embeddings produced by applying different dimensionality reduction methods. Sentence embeddings are created using pre-trained \texttt{stsb-bert-base} (uppermost), \texttt{stsb-bert-large} (supper-middle), \texttt{all-mpnet-base-v2} (lower-middle) and \texttt{sup-simcse-roberta-large} (lowermost) models. Results for the transductive and inductive settings are shown respectively on the left and right.}
\end{figure*}

\begin{figure*}[t!]
    \vspace{-0.5cm}
    \centering
    \subfigure[Transductive setting]{
        \rotatebox{90}{\footnotesize{~~~~~~~~~sup-simcse-roberta-large~~~~~~~~~~~~~paraphrase-xlm-r-multilingual-v1~~~~~~~~~~~~~msmarco-roberta-base-v2~~~~~~~~~~~~~~~~~~~~~~~all-mpnet-base-v2}}
        \begin{minipage}[b]{0.4\linewidth}
        \includegraphics[width=1.07\linewidth]{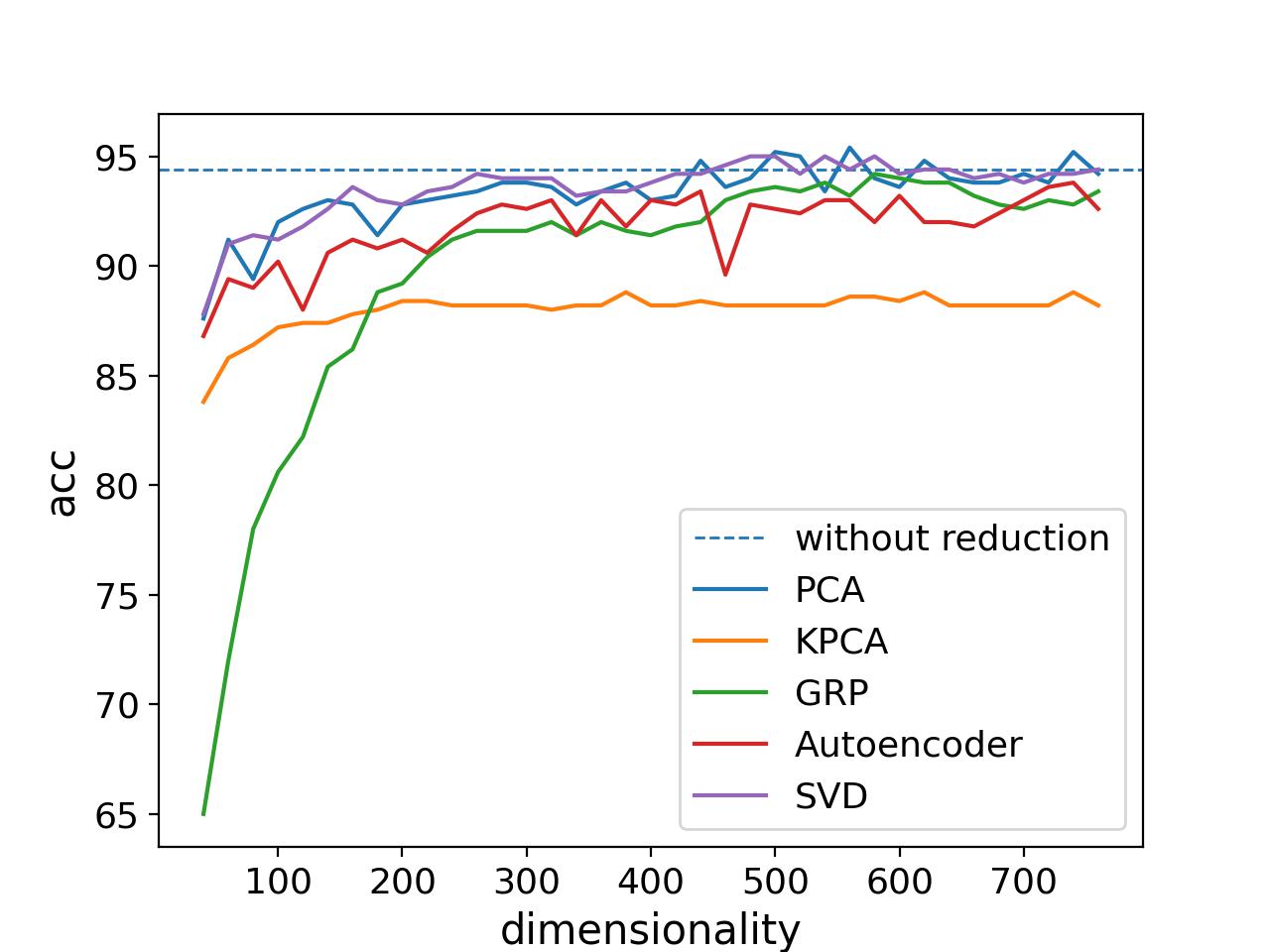}
        \vspace{0cm}
        \includegraphics[width=1.07\linewidth]{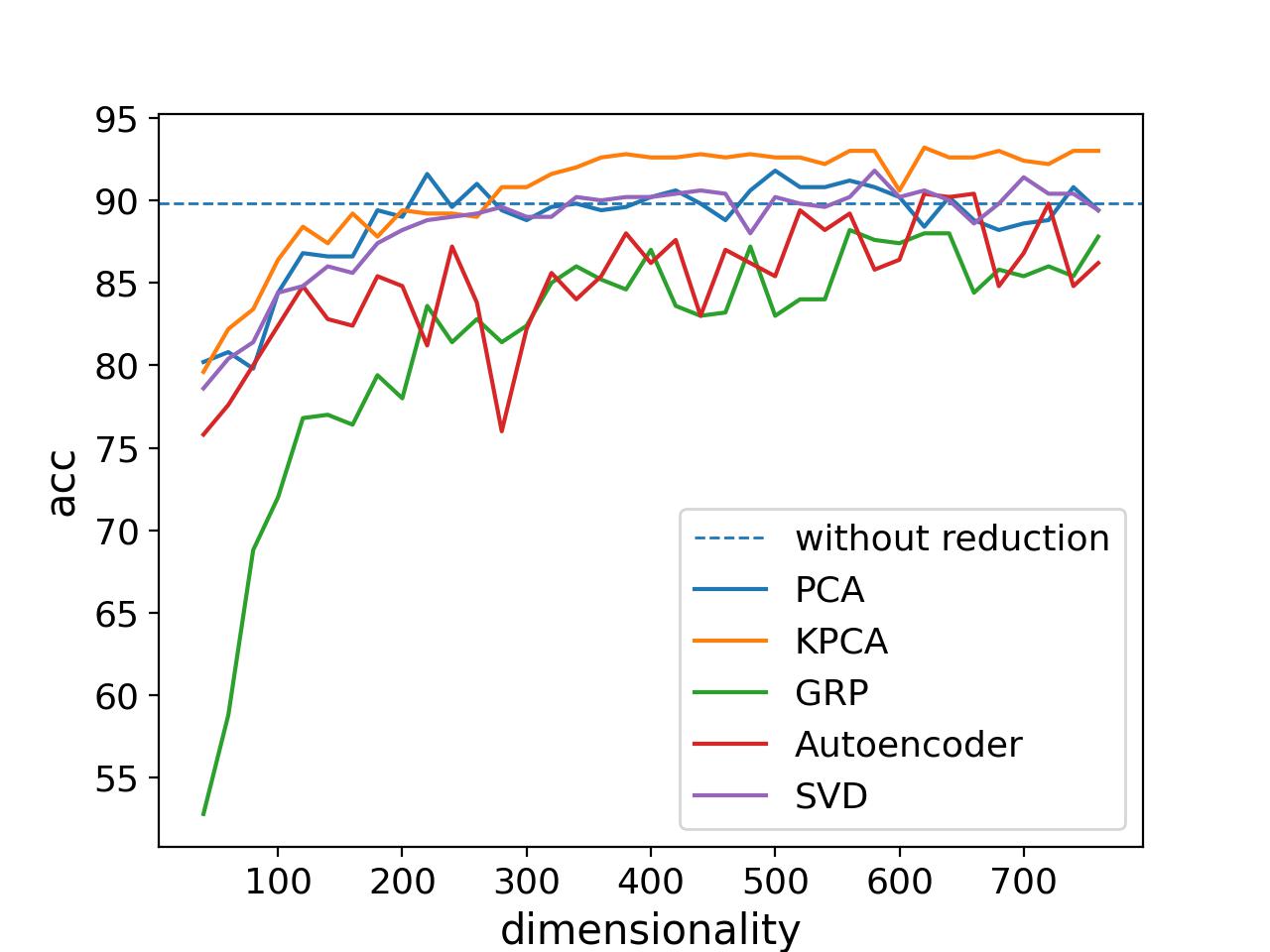}
        \vspace{0cm}
        \includegraphics[width=1.07\linewidth]{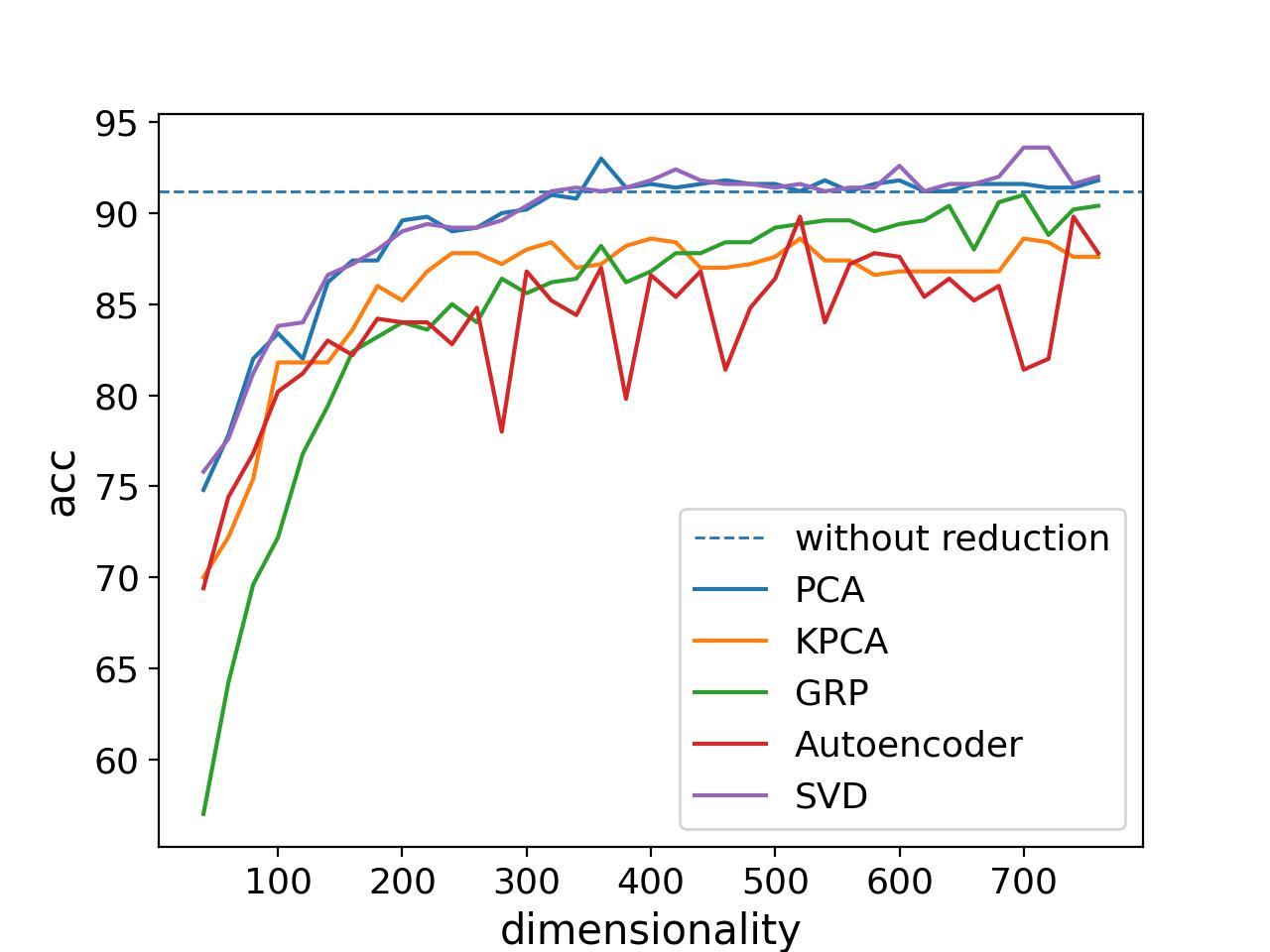}
        \vspace{0cm}
        \includegraphics[width=1.07\linewidth]{picture//TREC_trans_SimCSE.jpg}        
        \end{minipage}
    }
    \subfigure[Inductive setting]{
        \begin{minipage}[b]{0.4\linewidth} 
        \includegraphics[width=1.07\linewidth]{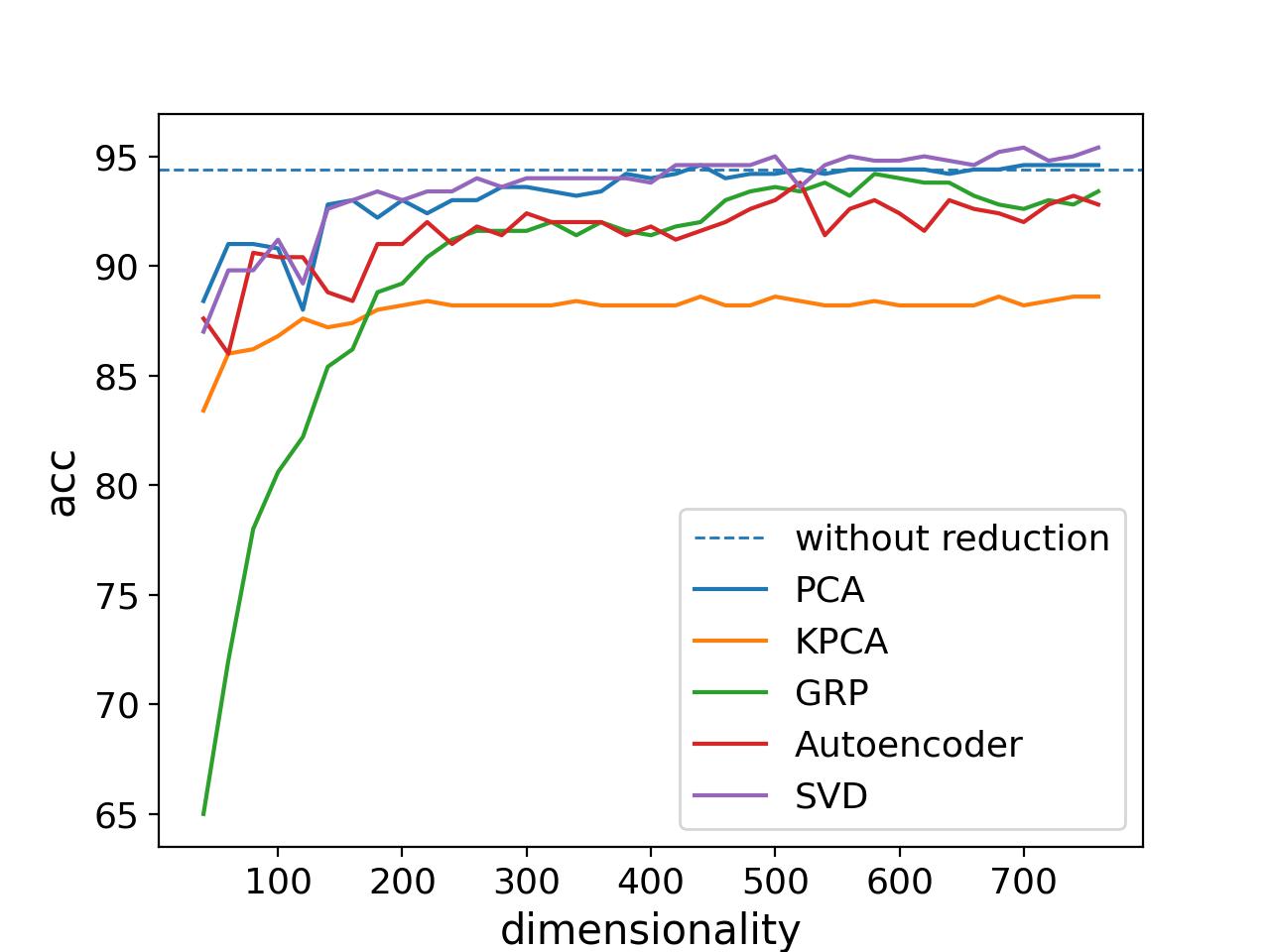}
        \vspace{0cm}
        \includegraphics[width=1.07\linewidth]{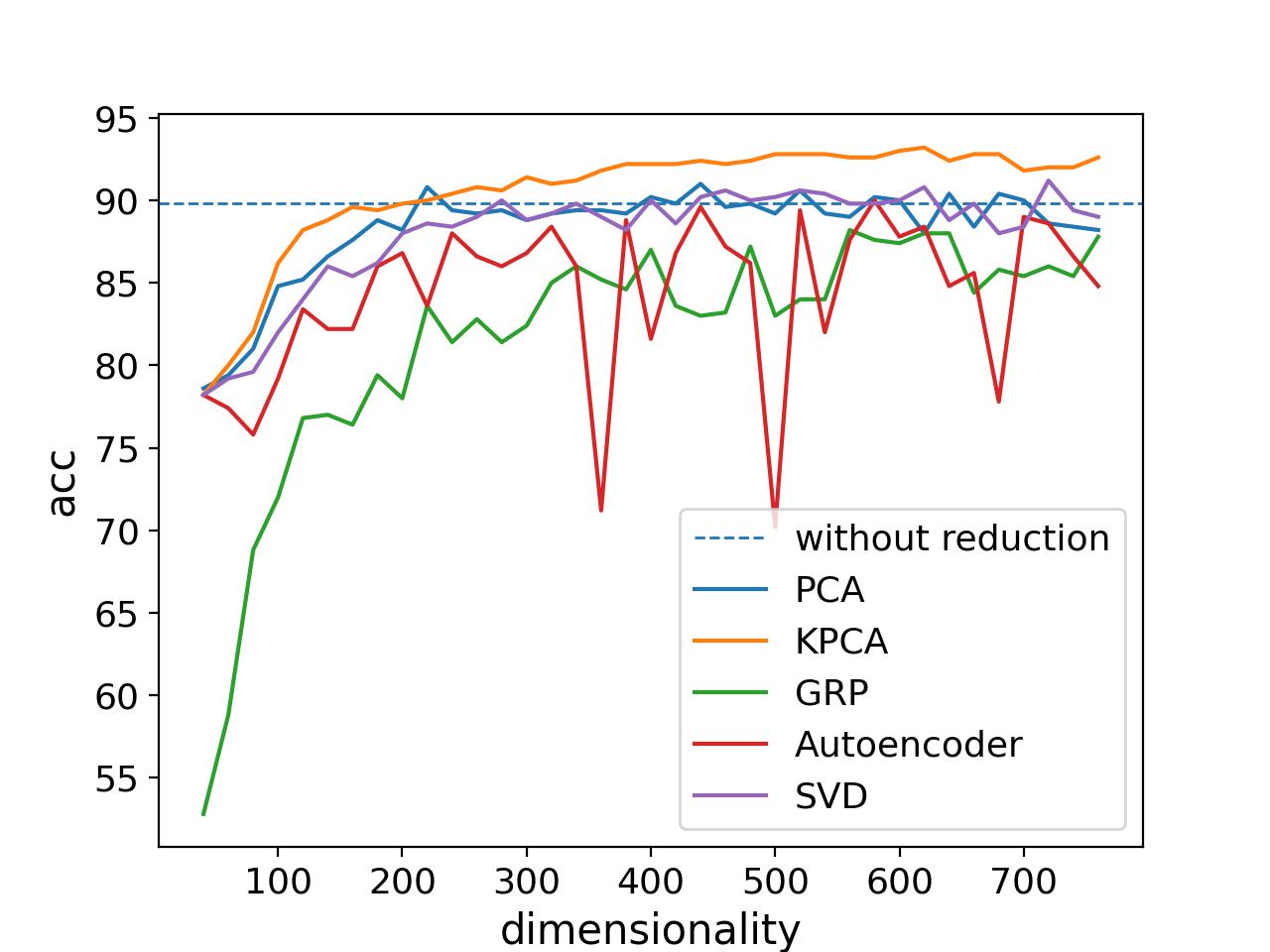}
        \vspace{0cm}
        \includegraphics[width=1.07\linewidth]{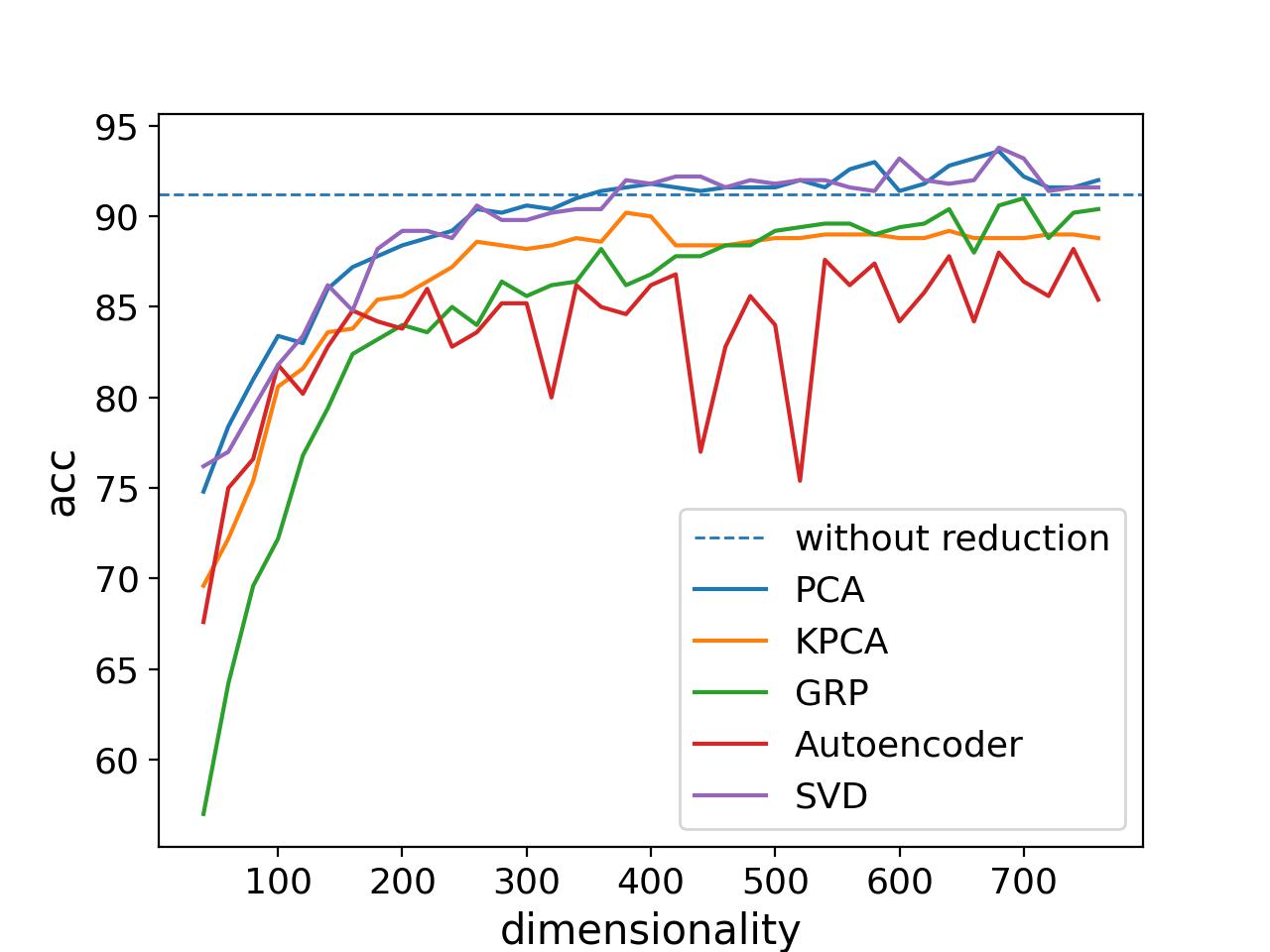}
        \vspace{0cm}
        \includegraphics[width=1.07\linewidth]{picture//TREC_in_SimCSE.jpg}
        \end{minipage}
    }
    \caption{Classification accuracy on TREC vs. the dimensionality of the sentence embeddings produced by applying different dimensionality reduction methods. Sentence embeddings are created using pre-trained \texttt{all-mpnet-base-v2} (uppermost), \texttt{msmarco-roberta-base-v2} (supper-middle), \texttt{paraphrase-xlm-r-multilingual-v1} (lower-middle) and \texttt{sup-simcse-roberta-large} (lowermost) models. Results for the transductive and inductive settings are shown respectively on the left and right.}
\end{figure*}

\begin{figure*}[t!]
    \centering
    \subfigure[Transductive setting]{
        \rotatebox{90}{\footnotesize{~~~~~~~~~sup-simcse-roberta-large~~~~~~~~~~~~~~paraphrase-xlm-r-multilingual-v1~~~~~~~~~~~~msmarco-roberta-base-v2~~~~~~~~~~~~~~~~~~~~~~~all-mpnet-base-v2}}
        \begin{minipage}[b]{0.4\linewidth}
        \includegraphics[width=1.07\linewidth]{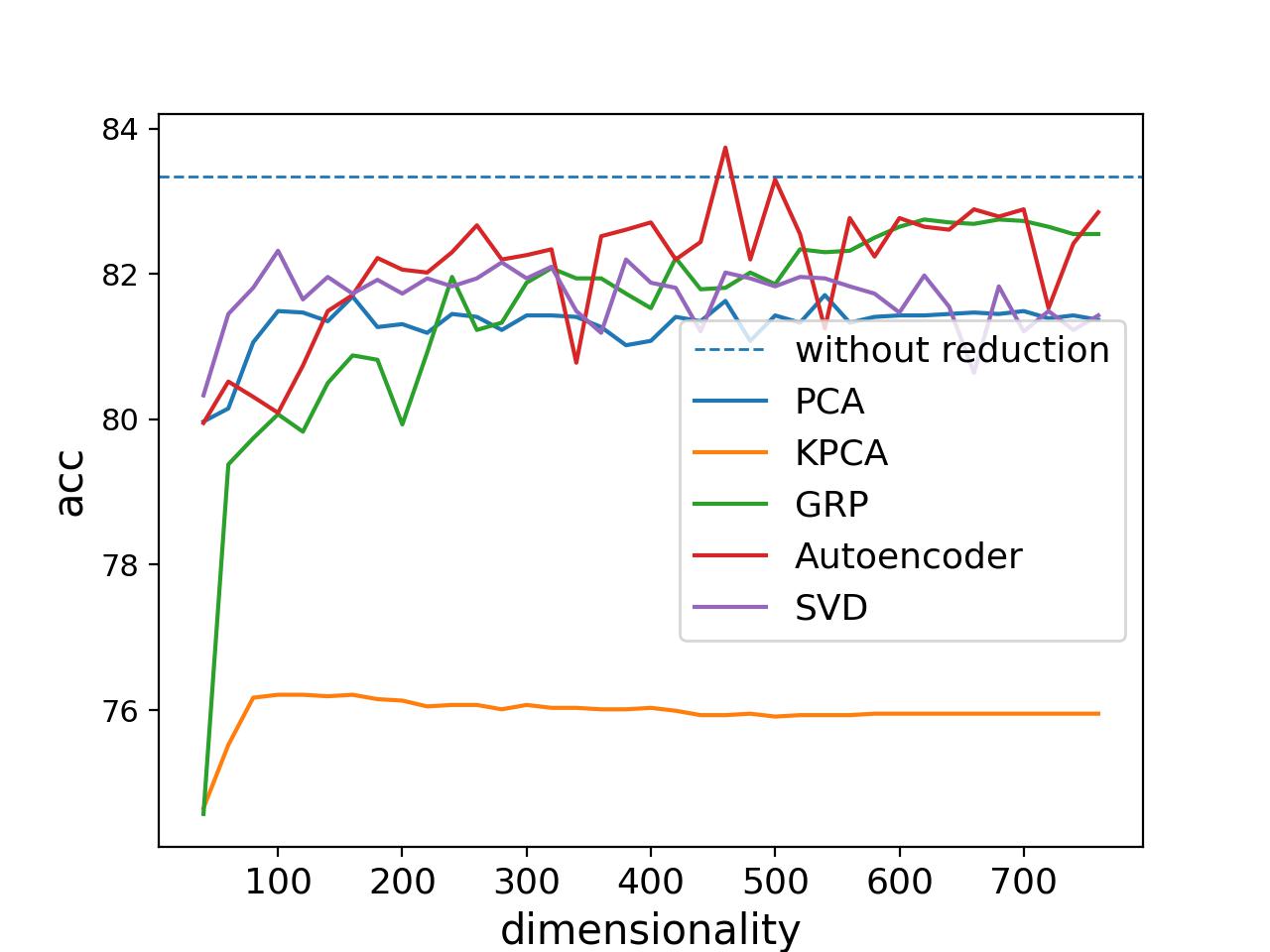}
        \vspace{0cm}
        \includegraphics[width=1.07\linewidth]{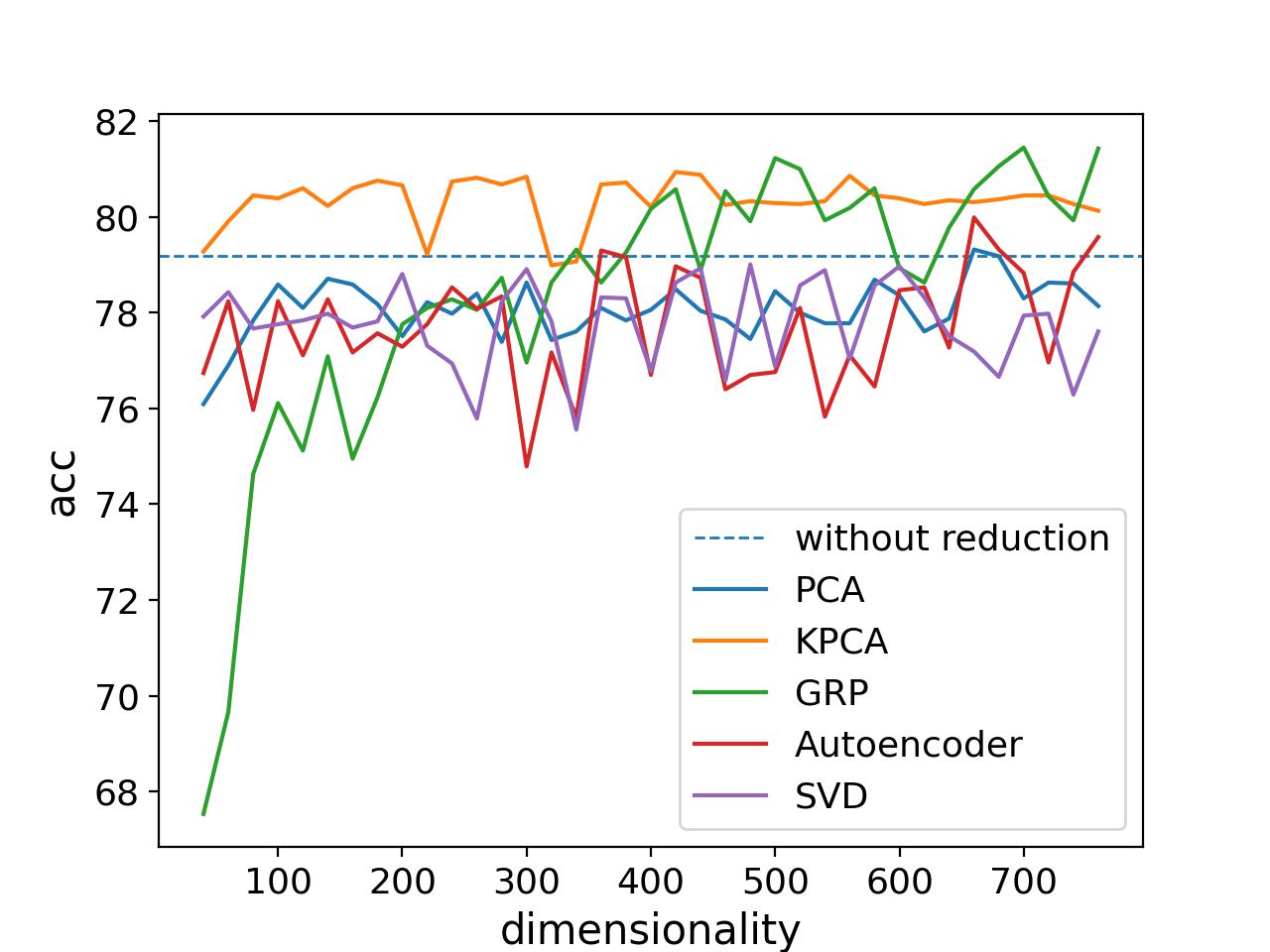}
        \vspace{0cm}
        \includegraphics[width=1.07\linewidth]{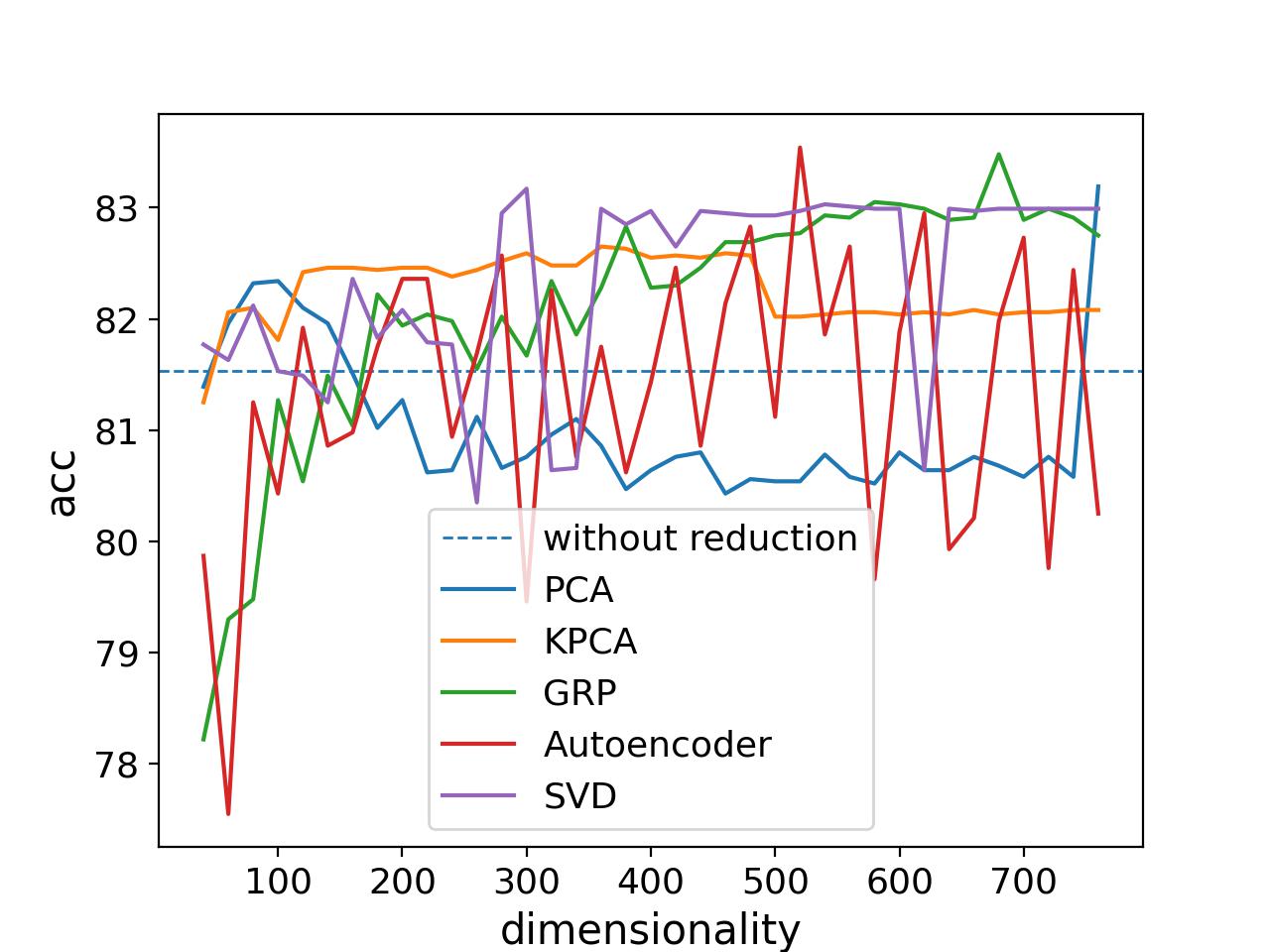}
        \vspace{0cm}
        \includegraphics[width=1.07\linewidth]{picture//SICKE_trans_SimCSE.jpg}
        
        \end{minipage}
    }
    \subfigure[Inductive setting]{
        \begin{minipage}[b]{0.4\linewidth} 
        \includegraphics[width=1.07\linewidth]{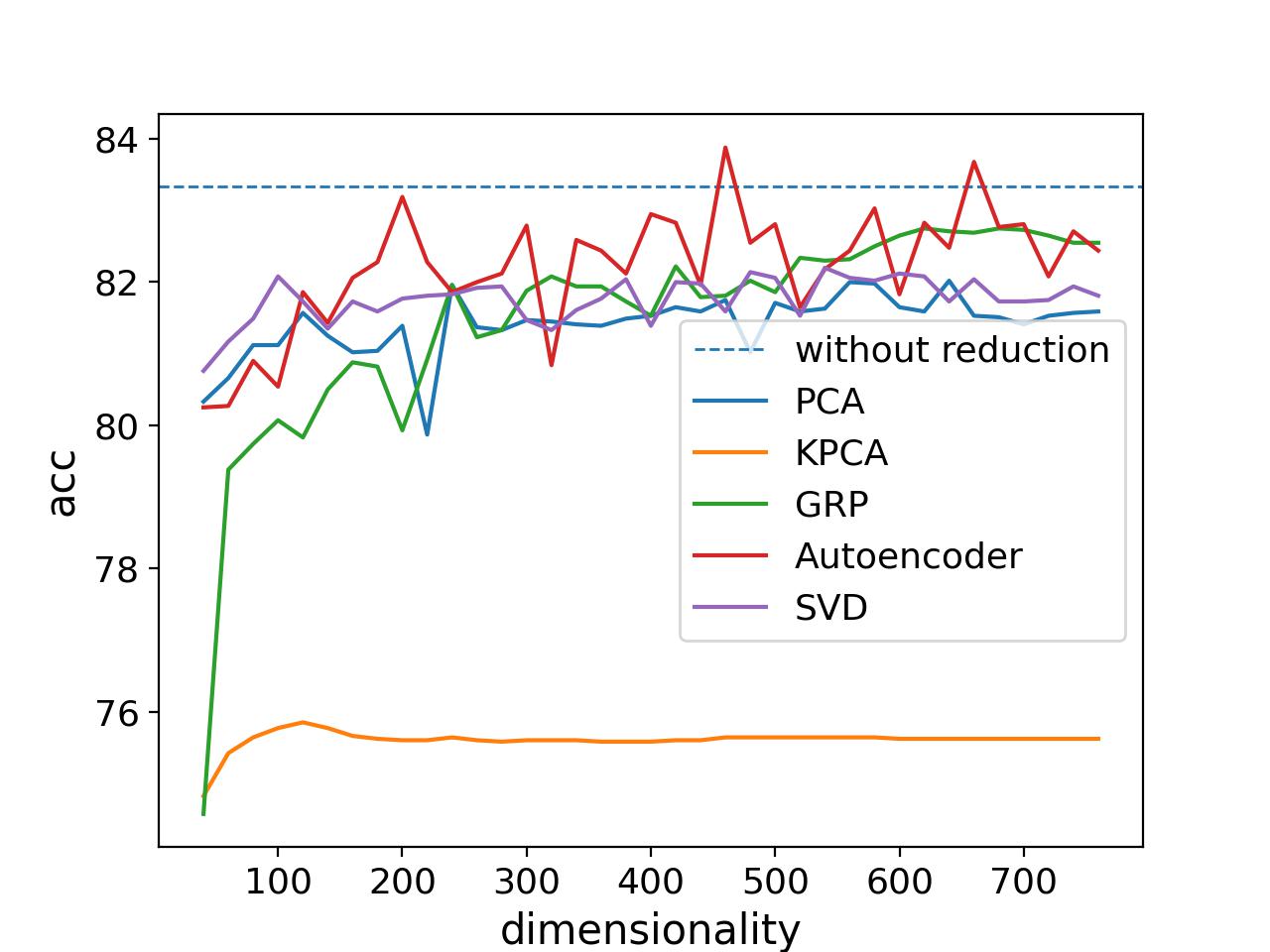}
        \vspace{0cm}
        \includegraphics[width=1.07\linewidth]{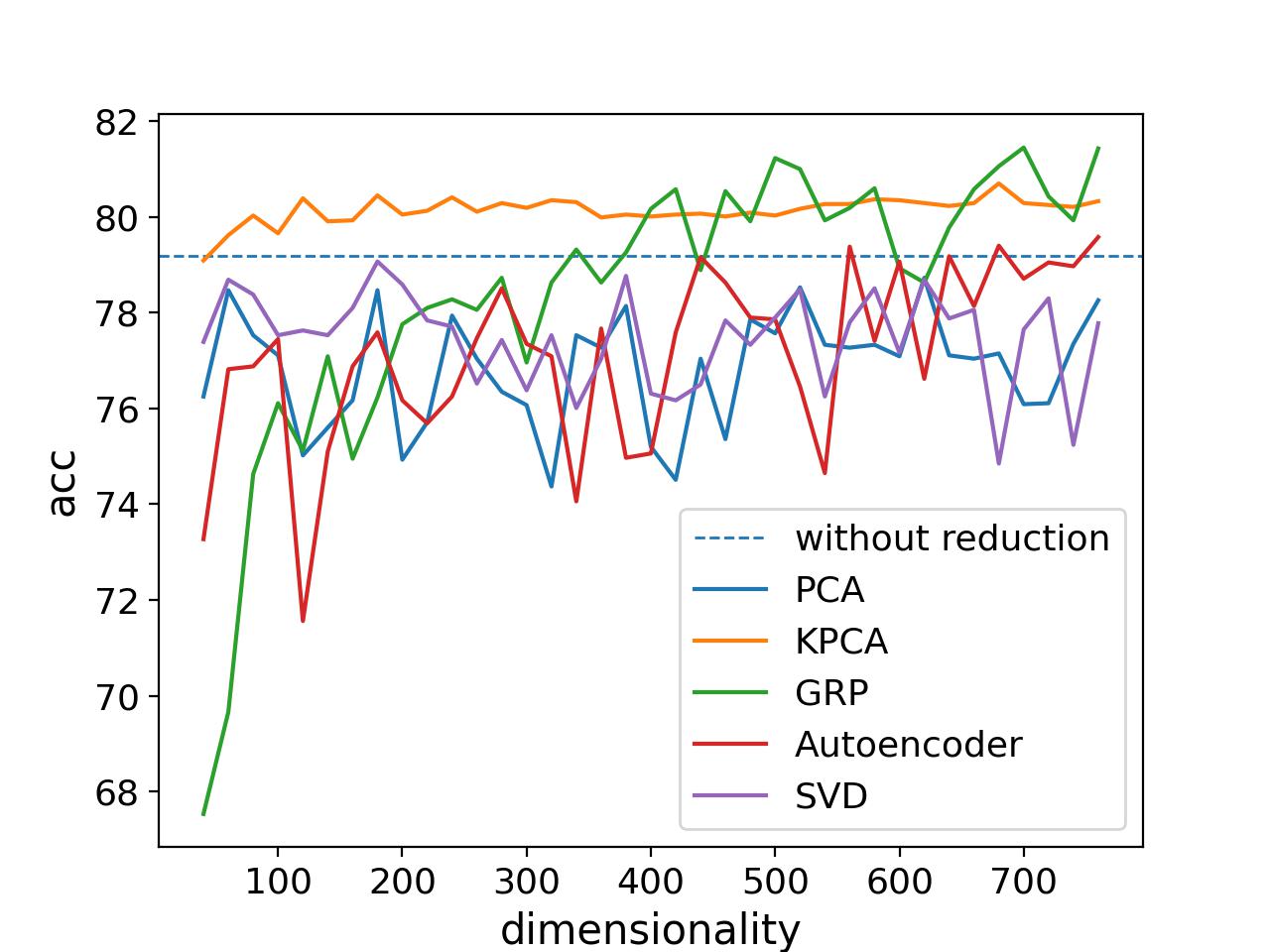}
        \vspace{0cm}
        \includegraphics[width=1.07\linewidth]{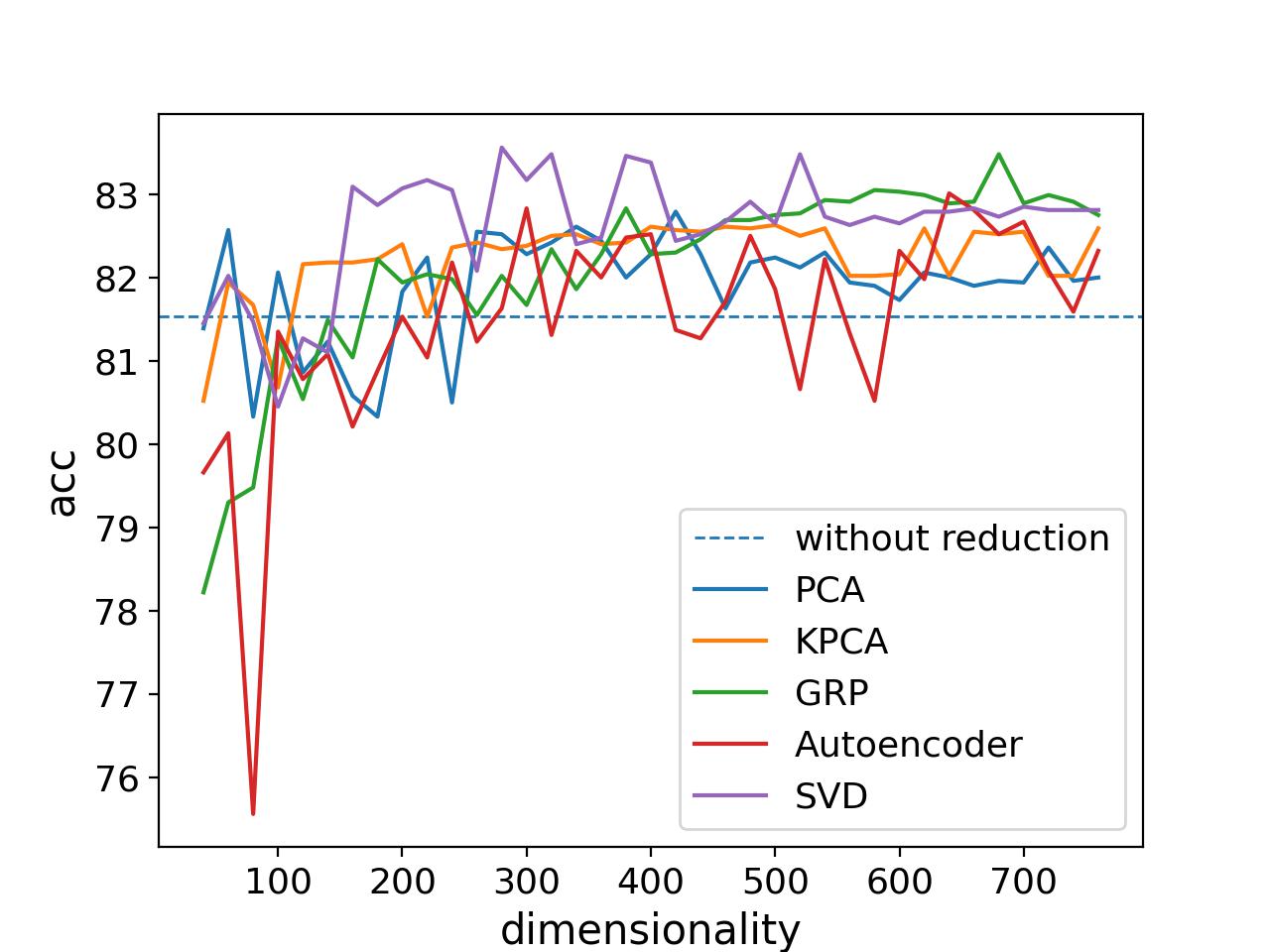}
        \vspace{0cm}
        \includegraphics[width=1.07\linewidth]{picture//SICKE_in_SimCSE.jpg}
        \end{minipage}
    }
    \caption{Accuracy on SICK-E vs. the dimensionality of the sentence embeddings produced by applying different dimensionality reduction methods. Sentence embeddings are created using pre-trained \texttt{all-mpnet-base-v2} (uppermost), \texttt{msmarco-roberta-base-v2} (supper-middle), \texttt{paraphrase-xlm-r-multilingual-v1} (lower-middle) and \texttt{sup-simcse-roberta-large} (lowermost) models. Results for the transductive and inductive settings are shown respectively on the left and right.}
\end{figure*}


 PCA outperforms uncompressed embeddings for some sentence encoders such as xlm-r on TREC. 
SVD performs comparably to PCA in most cases. However, SVD improves accuracy for both transductive and inductive settings for xlm-r on SICK-E. 
Meanwhile, PCA shows poor performance in the transductive setting in this case compared to the remainder of the methods.

KPCA shows varying levels of performance for different sentence encoders on different tasks. 
KPCA performs similarly to PCA in some situations, especially for mpnet and simcse on STS-B. 
KPCA reports excellent performance for some sentence encoders such as roberta on TREC and SICK-E and xlm-r on SICK-E, by even outperforming the original uncompressed embeddings. 

GRP reports suboptimal performance in many settings. 
In particular, mpnet on STS-B, GRP preserves much information in the original embeddings without a significant loss for dimensionalities over $200$, but shows a much lower performance than other methods. 
However, its performance on roberta and xlm-r for SICK-E is noteworthy.

Autoencoder gives relatively poor performance in many cases, especially for sentence encoders simcse, sbert-b, sbert-l and mpnet. 
Additionally, autoencoder has the most expensive train and infer costs among all the dimensionality reduction compared in this paper.

\end{document}